\documentclass[10pt,twocolumn,letterpaper]{article}

\usepackage{iccv}
\usepackage{times}
\usepackage{epsfig}
\usepackage{graphicx}
\usepackage{amsmath}
\usepackage{amssymb}
\usepackage{caption}

\usepackage{multirow}
\usepackage{subfigure} 
\usepackage{makecell}

\usepackage{amsmath,amsfonts,bm}

\def\Figref#1{Figure~\ref{#1}}

\def\Secref#1{Section~\ref{#1}}

\def\eqref#1{equation~\ref{#1}}
\def\Eqref#1{Equation~\ref{#1}}

\def\1{\bm{1}}

\def\vtheta{{\bm{\theta}}}

\def\vepsilon{{\bm{\epsilon}}}

\def\vc{{\bm{c}}}

\def\vf{{\bm{f}}}
\def\vg{{\bm{g}}}

\def\vv{{\bm{v}}}
\def\vw{{\bm{w}}}
\def\vx{{\bm{x}}}
\def\vy{{\bm{y}}}

\def\mF{{\bm{F}}}

\def\mH{{\bm{H}}}

\def\mS{{\bm{S}}}

\def\mU{{\bm{U}}}
\def\mV{{\bm{V}}}
\def\mW{{\bm{W}}}
\def\mX{{\bm{X}}}
\def\mY{{\bm{Y}}}

\DeclareMathAlphabet{\mathsfit}{\encodingdefault}{\sfdefault}{m}{sl}
\SetMathAlphabet{\mathsfit}{bold}{\encodingdefault}{\sfdefault}{bx}{n}

\def\sF{{\mathbb{F}}}

\def\sS{{\mathbb{S}}}

\def\sX{{\mathbb{X}}}

\newcommand{\R}{\mathbb{R}}

\newcommand{\normltwo}{L^2}
\newcommand{\normlp}{L^p}
\newcommand{\normmax}{L^\infty}

\DeclareMathOperator*{\argmin}{arg\,min}

\DeclareMathOperator{\sign}{sign}

\usepackage[pagebackref=true,breaklinks=true,letterpaper=true,colorlinks,bookmarks=false]{hyperref}

\iccvfinalcopy 

\def\iccvPaperID{5558} 

\begin{document}

\title{On Adversarial Robustness of 3D Point Cloud Classification under Adaptive Attacks}

\author{Jiachen Sun\\
University of Michigan\\
{\tt\small jiachens@umich.edu}
\and
Karl Koenig\\
University of Michigan\\
{\tt\small kamako@umich.edu}
\and
Yulong Cao\\
University of Michigan\\
{\tt\small yulongc@umich.edu}
\and
Qi Alfred Chen\\
University of California, Irvine\\
{\tt\small alfchen@uci.edu}
\and
Z. Morley Mao\\
University of Michigan\\
{\tt\small zmao@umich.edu}
}

\maketitle

\begin{abstract}
   3D point clouds play pivotal roles in various safety-critical applications, such as autonomous driving, which desires the underlying deep neural networks to be robust to adversarial perturbations. Though a few defenses against adversarial point cloud classification have been proposed, it remains unknown whether they are truly robust to adaptive attacks. To this end, we perform the first security analysis of state-of-the-art defenses and design adaptive evaluations on them. Our 100\% adaptive attack success rates show that current countermeasures are still vulnerable. Since adversarial training (AT) is believed as the most robust defense, we present the first in-depth study showing how AT behaves in point cloud classification and identify that the required symmetric function (pooling operation) is paramount to the 3D model's robustness under AT. Through our systematic analysis, we find that the default-used fixed pooling (\textit{e.g.,} \verb+MAX+ pooling) generally weakens AT's effectiveness in point cloud classification. Interestingly, we further discover that \textit{sorting-based} parametric pooling can significantly improve the models' robustness. Based on above insights, we propose \verb+DeepSym+, a deep symmetric pooling operation, to architecturally advance the robustness to 47.0\% under AT without sacrificing nominal accuracy, outperforming the original design and a strong baseline by 28.5\% ($\sim 2.6 \times$) and 6.5\%, respectively, in PointNet. 
\end{abstract}

\vspace{-0.8cm}
\section{Introduction}
\label{sec:intro}

Despite the prominent achievements that deep neural networks (DNN) have reached in the past decade, adversarial attacks~\cite{szegedy2013intriguing} are becoming the Achilles' heel in modern deep learning deployments, where adversaries generate imperceptible perturbations to mislead the DNN models. Numerous attacks have been deployed in various 2D vision tasks, such as classification~\cite{Carlini_2017}, object detection~\cite{song2018physical}, and segmentation~\cite{xie2017adversarial}. Since adversarial robustness is a critical feature, tremendous efforts have been devoted to defending against 2D adversarial images~\cite{guo2017countering,papernot2016distillation,madry2018towards}. However, Athalye \etal ~\cite{pmlr-v80-athalye18a} suggest that most of the current countermeasures essentially try to obfuscate gradients, which give a false sense of security. Besides, certified methods~\cite{zhang2019towards} often provide a lower bound of robustness, which are not helpful in real-world contexts. Adversarial training~\cite{madry2018towards}, thus, stands out to become the most effective and practical defense strategy since it well balances a strong guarantee with robustness to adaptive attacks.

\begin{figure}
\begin{center}
\includegraphics[width=\linewidth]{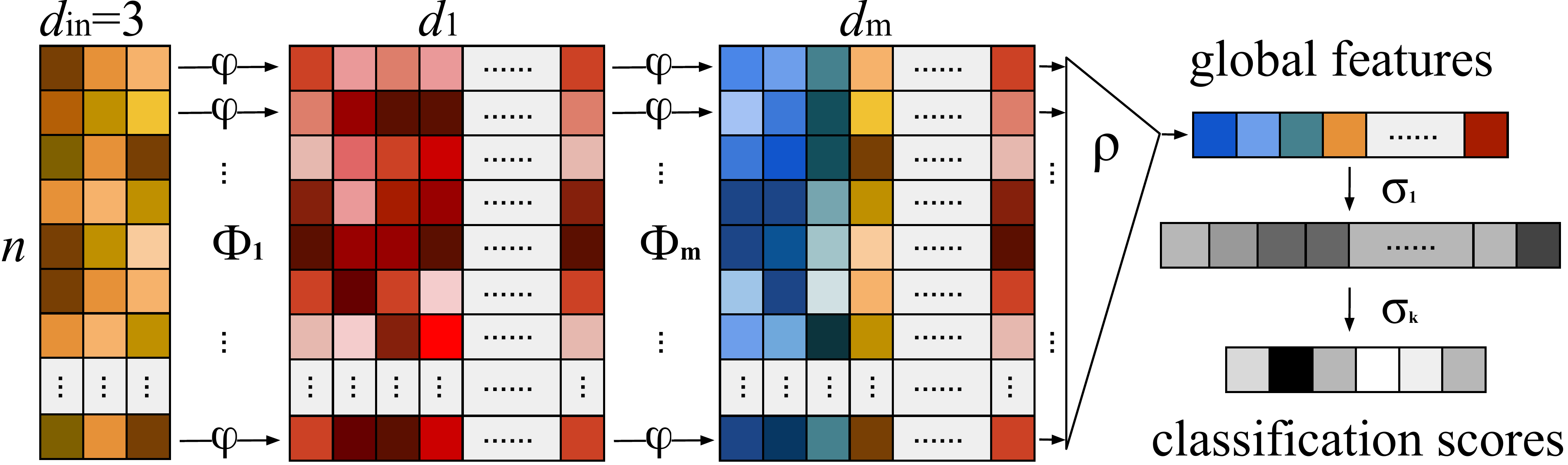}
\end{center}
\vspace{-0.15cm}
\caption{The general specification of point cloud classification $(\sigma \circ \rho \circ \Phi)(\sX)$, where $n$ is the number of points, $d_i$ is the number of hidden dimensions in the $i$-th feature map, $\Phi(\cdot)$ represents the permutation-equivariant layers, $\rho(\cdot)$ denotes the column-wise symmetric (permutation-invariant) function, and $\sigma(\cdot)$ is the fully connected layer.}
\label{fig:spec}
\vspace{-0.3cm}
\end{figure}

The emergence of 3D point cloud applications in safety-critical areas like autonomous driving raises public concerns about their security of DNN pipelines. A few studies~\cite{xiang2019generating,cao2019adversarial,255240} have demonstrated that various deep learning tasks on point clouds are indeed vulnerable to adversarial examples. Among them, classification is an essential and fundamental task on point clouds. While it seems intuitive to extend convolutional neural networks (CNN) from 2D to 3D for point cloud classification, it is actually not a trivial task. The difficulty mainly inherits from that point cloud is an \textit{sparse} and \textit{unordered} set structure that CNN cannot handle. Pioneering point cloud recognition models~\cite{qi2017pointnet,NIPS2017_6931} address this problem by leveraging a \textbf{symmetric function}, which is \textit{permutation-invariant} to the order of points, to aggregate local features, as illustrated in~\Figref{fig:spec}. Such a primitive has been universally adopted in many other complex learning tasks like semantic segmentation and object detection~\cite{lang2019pointpillars,yu2018pu}. Therefore, it is imperative to study and enhance its robustness to adversarial attacks.


Recently, a number of countermeasures have been proposed to defend against 3D adversarial point clouds. However, the failure of gradient obfuscation-based defenses in the 2D vision tasks motivates us to re-think whether current defense designs provide \textit{true} robustness~\cite{tramer2020adaptive} for 3D point cloud classification. Especially, DUP-Net~\cite{Zhou_2019_ICCV} and GvG-PointNet++~\cite{Dong_2020_CVPR} claim to improve the adversarial robustness significantly. However, we find that both defenses belong to gradient obfuscation through our analysis, hence further design white-box adaptive attacks to break their robustness. Our \textbf{100\%} attack success rates demonstrate that current defense designs can still be circumvented by adversaries.

As mentioned above, adversarial training (AT) is considered the most effective defense strategy; we thus perform the \textit{first} rigorous study of how AT behaves in point cloud classification by exploiting projected gradient descent (PGD) attacks~\cite{madry2018towards}. We identify that the default used \textbf{symmetric function} bottlenecks the effectiveness of AT. Specifically, popular models (\textit{e.g.,} PointNet) utilize fixed pooling operations like \verb+MAX+ and \verb+SUM+ pooling as their symmetric functions to aggregate features. Different from CNN-based models that usually apply pooling operations with a small sliding window (\textit{e.g.,} $2 \times 2$), point cloud classification models leverage such fixed pooling operations to aggregate features from a large number of candidates (\textit{e.g.,} 1024). We find that those fixed pooling operations inherently lack flexibility and learnability, which are not appreciated by AT. Moreover, recent research has also presented parametric pooling operations in set learning~\cite{wang2020softpoolnet,Zhang2020FSPool}, which also preserve permutation-invariance. We take a step further to systematically analyze point cloud classification models' robustness with parametric pooling operations under AT. Experimental results show that the \textit{sorting-based} pooling design benefits AT well, which vastly outperforms \verb+MAX+ pooling, for instance, in adversarial accuracy by 7.3\% without hurting the nominal accuracy\footnote{In this paper, we use nominal and adversarial accuracy to denote the model's accuracy on clean and adversarially perturbed data, respectively.}. 

Lastly, based on our experimental insights, we propose \verb+DeepSym+, a sorting-based pooling operation that employs deep learnable layers, to architecturally advance the adversarial robustness of point cloud classification under AT. \verb+DeepSym+ is intrinsically flexible and general by design. Experimental results show that \verb+DeepSym+ reaches the best adversarial accuracy in all chosen backbones, which on average, is a 10.8\% improvement compared to the default architectures.  
We also explore the limits of \verb+DeepSym+ based on PointNet due to its broad adoption in multiple 3D vision tasks~\cite{guo2020deep}. We obtain the best robustness on ModelNet40, which achieves the adversarial accuracy of 47.0\%, significantly outperforming the default \verb+MAX+ pooling design by 28.5\% ($\sim 2.6 \times$). In addition, we demonstrate that PointNet with \verb+DeepSym+ also reaches the best adversarial accuracy of 45.2\% under the most efficient AT on ModelNet10~\cite{Wu_2015_CVPR}, exceeding \verb+MAX+ pooling by 17.9\% ($\sim 1.7 \times$). 

Overall, our key contributions are summarized below:

\begin{minipage}[t]{0.95\linewidth}
\noindent $\bullet$ We design adaptive attacks on state-of-the-art point cloud defenses and achieve 100\% success rates.

\noindent $\bullet$ We conduct the first study to analyze how AT performs in point cloud recognition and identify sorting-based pooling can greatly benefit its robustness.

\noindent $\bullet$ We propose \verb+DeepSym+, a deep trainable pooling operation, to significantly improve the classification robustness by $\sim 2.6 \times$ compared to the baseline methods.
\end{minipage}

\vspace{-0.2cm}
\section{Background and Related Work}
\label{sec:bg}

\textbf{3D point cloud classification.} Early works attempt to classify point clouds by adapting deep learning models in the 2D space~\cite{su2015multi,yu2018multi}. PointCNN~\cite{li2018pointcnn} tries to address the unorderness problem by learning a permutation matrix, which is, however, still non-deterministic. DeepSets~\cite{NIPS2017_6931} and PointNet~\cite{qi2017pointnet} are the first to achieve end-to-end learning on point cloud classification and formulate a general specification (\Figref{fig:spec}) for point cloud learning. PointNet++~\cite{qi2017pointnet++} and DGCNN~\cite{dgcnn} build upon PointNet set abstraction to better learn local features by exploiting \textit{k}-nearest neighbors. Lately, DSS~\cite{maron2020learning} generalizes DeepSets to enable complex functions in set learning. Besides, ModelNet40~\cite{Wu_2015_CVPR} is the most popular dataset for benchmarking point cloud classification, which consists of 12,311 CAD models belonging to 40 categories. The numerical range of the point cloud data is normalized to $[-1,1]$ in ModelNet40. 

\textbf{Adversarial Machine Learning.} DNN models are principally shown vulnerable to adversarial attacks~\cite{Carlini_2017,goodfellow2014explaining,xie2017adversarial}. Though there are various methods to create adversarial examples, projected gradient descent (PGD)-based attacks~\cite{madry2018towards} are the most recognized:
\vspace{-0.2cm}
\begin{equation}
    \mX_{t+1} = \Pi_{\mX + \mathcal{S}}(\mX_{t} + \alpha \cdot \mathrm{normalize}(\nabla_{\mX_{t}} \mathcal{L}(\mX_t,\vtheta,\vy)))
    \label{eq:pgd}
\end{equation}
where $\mX_t$ is the adversarial example in the $t$-th attack iteration, $\Pi$ is the projection function to project the adversarial example to the pre-defined perturbation space $\mathcal{S}$, and $\alpha$ is the step size. More details are provided in Appendix A.1.

\textbf{Adversarial attacks and defenses on point clouds.} Xiang~\etal~\cite{xiang2019generating} perform the first study to extend C\&W attack~\cite{Carlini_2017} to point cloud classification. Wen~\etal~\cite{wen2019geometry} improve the loss function in C\&W attack to realize attacks with smaller perturbations and ~\cite{hamdi2019advpc} present black-box attacks on point cloud classification. Recently,~\cite{Zhou_2019_ICCV} and~\cite{Dong_2020_CVPR} propose to defend against adversarial point clouds by input transformation and adversarial detection. Besides,~\cite{liu2019extending} conduct a preliminary investigation on extending countermeasures in the 2D space to defend against na\"ive attacks like FGSM~\cite{goodfellow2014explaining} on point cloud data. Liu~\etal~\cite{liu2021pointguard} propose to certify the robustness of point cloud recognition with a threat model only considering the number of modified points. In this work, we first design adaptive attacks to break existing defenses and analyze the adversarial robustness of point cloud classification under adversarial training constrained by widely recognized $\normlp$ norms. 

\vspace{-0.2cm}
\section{Breaking Existing Point Cloud Defenses}
\label{sec:attack}
In this section, we introduce the design and evaluation of our adaptive attacks on state-of-the-art defenses, DUP-Net~\cite{Zhou_2019_ICCV} and GvG-PointNet++~\cite{Dong_2020_CVPR}, followed by the suggestions of Carlini~\etal~\cite{tramer2020adaptive}.

\begin{figure}[t]
\vspace{-0.1cm}

\vspace{-0.02cm}
\begin{center}
\includegraphics[width=\linewidth]{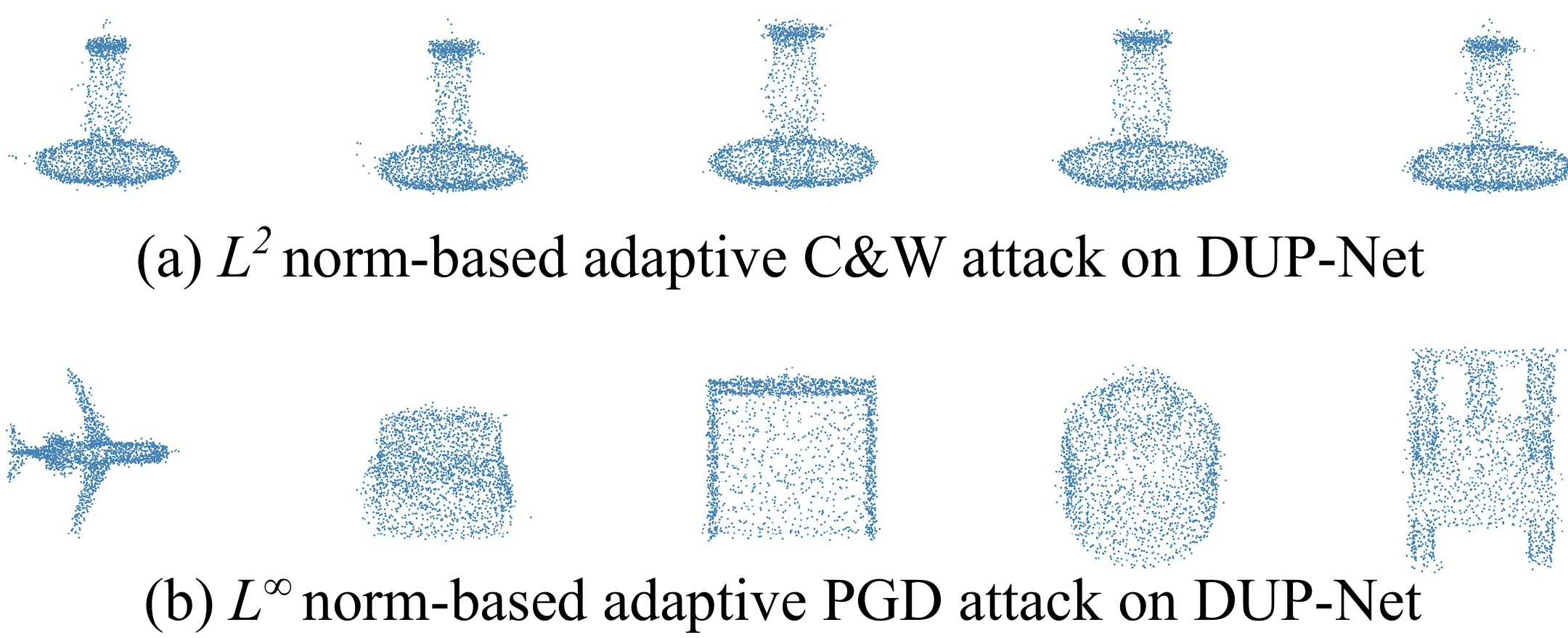} 
\end{center}

\vspace{-0.5cm}
\begin{center}
\includegraphics[width=\linewidth]{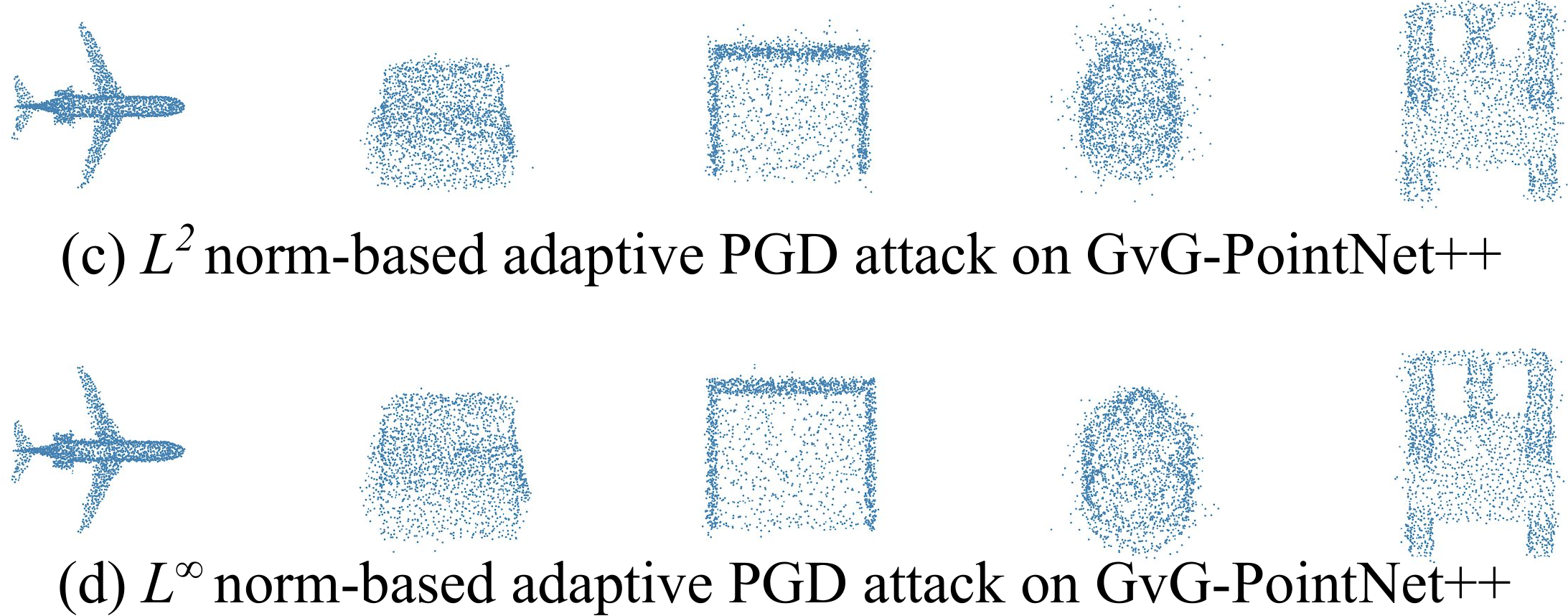} 
\end{center}
\vspace{-0.3cm}
\caption{Sampled visualizations of adversarial examples generated by adaptive attacks ($\epsilon = 0.05$ and $\delta = 0.16$). More visualizations can be found in Appendix A.2.}
\vspace{-0.3cm}
\end{figure}

\subsection{Adaptive Attacks on DUP-Net}
\label{sec:attack_dup}
\textbf{DUP-Net} (ICCV'19) presents a denoiser layer and upsampler network structure to defend against adversarial point cloud classification. The denoiser layer $g: \sX \rightarrow \sX'$ leverages \textit{k}NN (\textit{k}-nearest neighbor) for outlier removal. Specifically, the \textit{k}NN of each point $\vx_i$ in point cloud $\sX$ is defined as $knn(\vx_i,k)$ so that the average distance $d_i$ of each point $\vx_i$ to its \textit{k}NN is denoted as:
\vspace{-0.1cm}
\begin{equation}
d_i = \frac{1}{k} \sum_{\vx_j \in knn(\vx_i,k)} ||\vx_i - \vx_j||_2 \;, \quad i = \{1, 2, \dots, n \}
\label{eq:knn}
\end{equation}
where $n$ is the number of points. The mean $\mu = \frac{1}{n} \sum_{i=1}^{n} d_i$ and standard deviation $\sigma = \sqrt{\frac{1}{n} \sum_{i=1}^{n}(d_i - \mu)^2}$ of all these distances are computed to determine a distance threshold as $\mu + \alpha \cdot \sigma$ to trim the point clouds, where $\alpha$ is a hyper-parameter. As a result, the denoised point cloud is represented as $\sX' = \{\vx_i \;|\; d_i < \mu + \alpha \cdot \sigma\}$.
The denoised point cloud $\sX'$ will be further fed into PU-Net~\cite{yu2018pu}, defined as $p: \sX' \rightarrow \sX''$, to upsample $\sX'$ to a fixed number of points. Combined with the classifier $f$, the integrated DUP-Net can be noted as $(f\circ p\circ g)(\sX)$. The underlying hypothesis is that the denoiser layer will eliminate the adversarial perturbations and the upsampler network will re-project the denoised off-manifold point cloud to the natural manifold.

\textbf{Analysis.} The upsampler network $p$ (\textit{i.e.,} PU-Net) is differentiable and can be integrated with the classification network $f$. Therefore, $f \circ p$ is clearly vulnerable to gradient-based adaptive attacks. Although the denoiser layer $g$ is not differentiable, it can be treated as \textit{deterministic masking}: $\mathcal{M}(\vx_i) = \1_{d_i < \mu + \alpha \cdot \sigma}$ so that the gradients can still flow through the masked points. By involving $\mathcal{M}(\vx_i)$ into the iterative adversarial optimization process: 
\vspace{-0.1cm}
\begin{equation}
\nabla_{\vx_i}(f\circ p\circ g)(\sX) |_{\vx_i = \hat{\vx}} \approx \nabla_{\vx_i}(f\circ p)(\sX) |_{\vx_i = \hat{\vx}\cdot\mathcal{M}(\hat{\vx})}
\end{equation}
\noindent similar to Backward Pass Differentiable Approximation (BPDA)~\cite{pmlr-v80-athalye18a}, attackers may still find adversarial examples.

\textbf{Experimentation.}  We leverage the official codebase\footnote{\url{https://github.com/RyanHangZhou/DUP-Net}} of DUP-Net for experimentation. Specifically, a PointNet~\cite{qi2017pointnet} trained on ModelNet40 is used as the classifier $f$. For the PU-Net, the upsampled number of points is 2048, and the upsampling ratio is 2. For the adaptive attacks, we exploit targeted $\normltwo$ norm-based C\&W attack and untargeted $\normmax$ norm-based PGD attack with 200 iterations (PGD-200). Detailed setups are elaborated in Appendix A.1. 

\begin{table}[t]
\renewcommand\arraystretch{1.1}
\setlength\tabcolsep{3pt}
\footnotesize
\caption{Adversarial accuracy under adaptive attacks on PU-Net and DUP-Net. For the denoiser layer $g$, $k=2$ and $\alpha = 1.1$ are set the same as~\cite{Zhou_2019_ICCV}. $\dagger$ denotes the attack setups evaluated in the original DUP-Net paper~\cite{Zhou_2019_ICCV}.}
\vspace{-0.3cm}
\label{tb:adaptive_1}
\begin{center}
\begin{tabular}{|c|c|c|c|c|}
\hline
\multirow{2}*{\makecell*[c]{\bf Attack Method}}  &\multicolumn{3}{c|}{\bf Adversarial Accuracy}
&\multirow{2}*{\makecell[c]{\bf Mean $\normltwo$ \\ \bf Norm \\ \bf Distance} }\\
\cline{2-4}
&PointNet ($f$) & \makecell[c]{PU-Net \\ ($f \circ p$)} & \makecell[c]{DUP-Net \\ ($f \circ p \circ g$)} &   \\
\hline
\hline
Clean point cloud    &88.3\%     &87.5\%    &86.3\%     &0.0 \\
\hline
C\&W attack on $f$ $\dagger$   &\textbf{0.0\%}     &23.9\%    &84.5\%     &0.77 \\
\hline
C\&W attack on $f \circ p$  &2.3\%   &\textbf{0.0\%}    &74.7\%        &0.71 \\
\hline
\makecell[c]{Adaptive attack \\ on $f \circ p \circ g$}  &1.1\%   &0.8\% &\textbf{0.0\%}     &1.62 \\
\hline
PGD attack ($\epsilon = 0.01$) &7.1\%   &5.9\% &5.4\%     &- \\
\hline
PGD attack ($\epsilon = 0.025$)  &3.5\%   &2.8\% &2.1\%     &- \\
\hline
PGD attack ($\epsilon = 0.05$)  &1.3\%   &1.0\% &0.8\%     &- \\
\hline
PGD attack ($\epsilon = 0.075$)  &\textbf{0.0\%}   &\textbf{0.0\%} &\textbf{0.0\%}     &- \\
\hline
\end{tabular}
\end{center}
\vspace{-0.6cm}
\end{table}

\begin{figure*}[!t]
\vspace{-0.15cm}
\begin{minipage}[t]{0.64\linewidth}
\vspace{-0.1cm}
\setlength\tabcolsep{3pt}
\footnotesize
\captionof{table}{Adversarial accuracy under $\normlp$ norm-based adaptive attacks on GvG-PointNet++. $\epsilon$ and $\delta$ are the perturbation boundaries. $\dagger$ denotes the attack setups evaluated in the original GvG-PointNet++ paper~\cite{Dong_2020_CVPR}.}
\label{tb:adaptive_2}
\vspace{-0.6cm}
\begin{center}
\begin{tabular}{|c|c|c|c|c|c|c|c|}
\hline
\multirow{2}*{\makecell*[c]{\bf Target Loss}}  &\multicolumn{4}{c|}{\bf Adversarial Accuracy ($\normmax$)} &\multicolumn{3}{c|}{\bf Adversarial Accuracy ($\normltwo$)}\\
\cline{2-8}
&$\epsilon = 0.01$ &$\epsilon = 0.025$ &$\epsilon = 0.05$ &$\epsilon = 0.075$   &$\delta = 0.08$ &$\delta = 0.16$ &$\delta = 0.32$ \\
\hline
\hline
\makecell[c]{$\mathcal{L}_{xent}$$\dagger$}  &30.6\%   &21.4\%  &5.4\%     &1.8\% &25.2\%   &16.9\%  &15.4\% \\
\hline
$\mathcal{L}_{adv}$  &20.1\%   &12.6\%  &2.2\%     &\textbf{0.0\%} &\textbf{7.5\%}   &4.4\%  &\textbf{2.1\%}\\
\hline
$\mathcal{L}_{gather}$   &\textbf{17.9\%}   &\textbf{8.1\%} &\textbf{0.0\%}     &\textbf{0.0\%} &8.5\%   &\textbf{4.1\%} &2.7\%\\
\hline
\end{tabular}
\end{center}

\setlength\tabcolsep{4pt}
\footnotesize
\caption{Adversarial robustness of models with fixed pooling operations under PGD-200 at $\epsilon=0.05$.}
\vspace{-0.3cm}
\label{tb:fixed_adv_train}
\begin{center}
\begin{tabular}{|c|c|c|c|c|c|c|}
\hline
\multirow{2}*{\makecell[t]{\bf Pooling Operation}} &\multicolumn{3}{c|}{\bf Nominal Accuracy} &\multicolumn{3}{c|}{\bf Adversarial Accuracy} \\
\cline{2-7}
&PointNet &DeepSets  &DSS  &PointNet &DeepSets  &DSS  \\
\hline
\hline
\verb+MAX+ &80.5\%   &71.1\%  &78.8\% &16.1\% &21.8\% &21.5\%\\
\hline
\verb+SUM+  &76.3\%   &54.1\%  &73.3\% &\textbf{25.1\%} &\textbf{24.8\%} &\textbf{25.3\%}\\
\hline
\verb+MEDIAN+   &\textbf{84.6\%}   &\textbf{72.7\%} &\textbf{82.4\%} &7.5\% &11.0\% &9.3\%\\
\hline
\end{tabular}
\end{center}
\vspace{-0.3cm}

\end{minipage}
\hfill
\begin{minipage}[t]{0.33\linewidth}
\vfill
\begin{center}
\vspace{2pt}
\includegraphics[width=\linewidth]{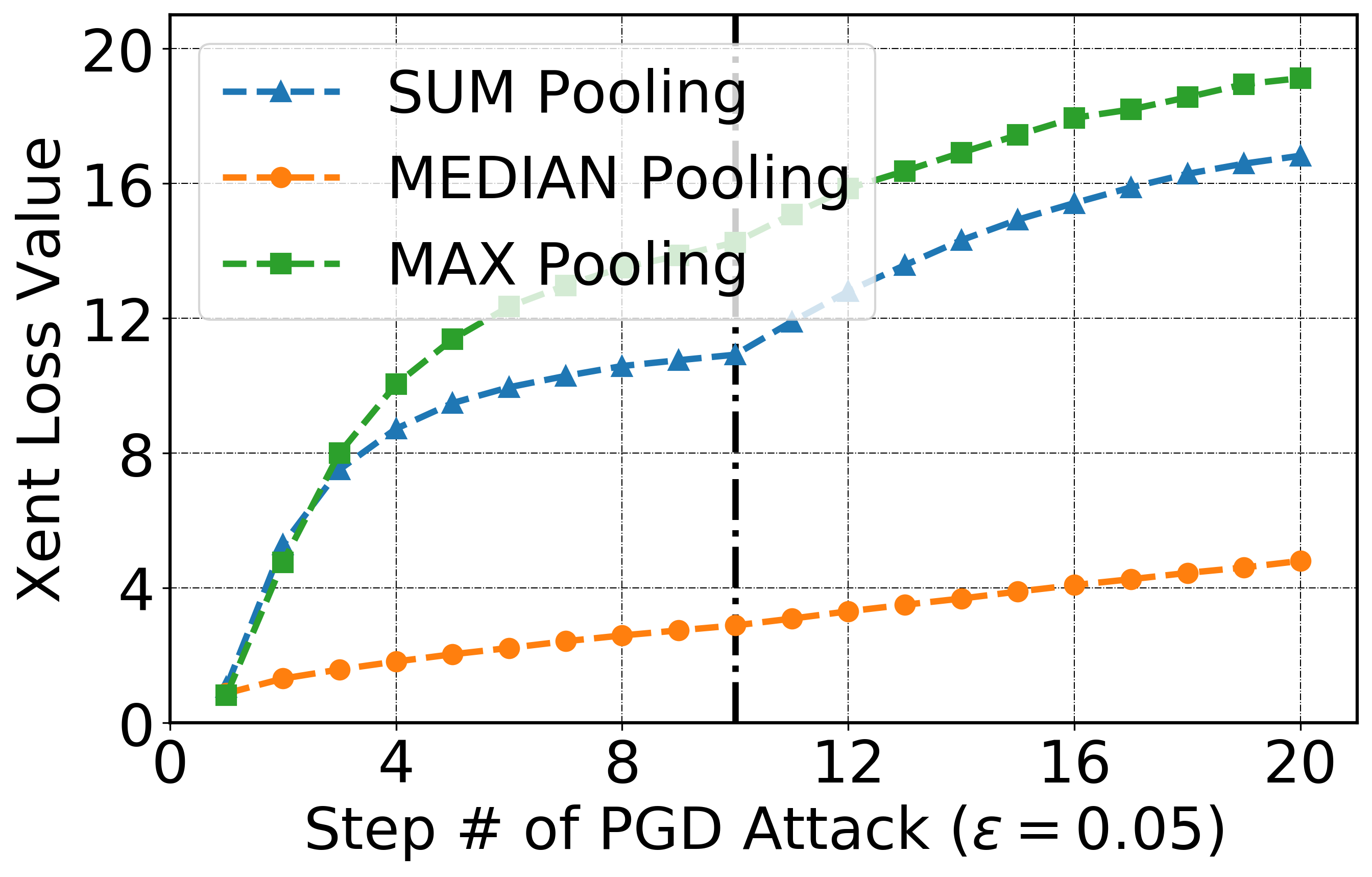} 
\end{center}
\caption{Xent loss of PGD attack on PointNet with three fixed pooling operations (each value is averaged over 100 runs from random starting points).}
\label{fig:loss}
\end{minipage}
\vspace{-0.3cm}
\end{figure*}

\textbf{Discussion.} As presented in Table~\ref{tb:adaptive_1}, adaptive C\&W attacks achieve 100\% success rates on DUP-Net. Though the mean distance of adversarial examples targeting DUP-Net is larger than those targeting PU-Net, they are almost indistinguishable by human perception, as visualized in Appendix A.2. We find that na\"ive PGD attacks are also effective since the upsampler network is sensitive to $\normmax$ norm-based perturbations, which also showcase the fragility of the defense pipeline. The design of DUP-Net is similar to ME-Net~\cite{yang2019me} in the 2D space, which recently has been shown vulnerable to adaptive attacks~\cite{tramer2020adaptive}. We hereby demonstrate that such input transformation-based defenses cannot offer true robustness to point cloud classification, either.




\vspace{-0.1cm}
\subsection{Adaptive Attacks on GvG-PointNet++}
\label{sec:attack_gvg}

\textbf{GvG-PointNet++} (CVPR'20) introduces gather vectors in the 3D point clouds as an adversarial indicator to detect adversarial perturbations. The original PointNet++~\cite{qi2017pointnet++} aggregates local features $\vf_i$ hierarchically to make final classification. Gather vectors $\vv_i$ are learned from local features $\vf_i$ to indicate the global center $\vc_i$ of a point cloud sample. If the indicated global center $\vc_i$ is far away from the ground-truth global center $\vc_g$, the corresponding local feature $\vf_i$ will be masked out:
\begin{equation}
\vc_i = \vx_{ci} + \vv_i \;; \quad \mathcal{M}_i = \1_{|| \vc_g - \vc_i || < r} \;; \quad \sF_g = \{\vf_i \cdot \mathcal{M}_i\}
\label{eq:gvg}
\end{equation}
where $\vx_{ci}$ is the geometry center of the local point set, $r$ is the distance threshold to mask the local feature, and $\sF_g$ is the cleaned feature set for final classification. To train GvG-PointNet++, it is necessary to optimize an auxiliary loss to correctly learn the gather vectors besides the default cross-entropy (xent) loss:
\vspace{-0.2cm}
\begin{equation}
\mathcal{L}_{total} = \mathcal{L}_{xent} + \lambda \cdot \mathcal{L}_{gather}\;,\quad \mathcal{L}_{gather} = \sum_{i=1}^{n'} || \vc_i - \vc_g ||_1
\label{eq:gvg_loss}
\end{equation}
where $n'$ is the number of local features and $\lambda$ is a hyper-parameter. Thus, GvG-PointNet++ essentially applies self-attention to the local feature learning and relies on it for potential robustness improvement. 

\textbf{Analysis.} Dong~\etal~\cite{Dong_2020_CVPR} evaluate white-box adversaries on GvG-PointNet++ with na\"ive $\normltwo$ norm-based PGD attacks. Specifically, only $\mathcal{L}_{xent}$ is utilized as the objective loss in the adversarial optimization process so that the masking $\mathcal{M}_i$ will hinder the gradient propagation. However, since $\mathcal{M}_i$ is learned from the network itself, it is highly possible to further break this obfuscation with $\mathcal{L}_{gather}$ considered. The adaptive attack can be then formulated as an optimization problem with the loss function:
\vspace{-0.1cm}
\begin{equation}
\mathcal{L}_{adv} = \mathcal{L}_{xent} - \beta \cdot \mathcal{L}_{gather}
\label{eq:gvg_adap_loss}
\end{equation}
where $\beta$ is a tunable hyper-parameter. By maximizing $\mathcal{L}_{adv}$ with $\normltwo$ norm-based PGD attacks, adversaries not only strive to enlarge the adversarial effect but also minimize the perturbations on gather vectors. We also find that GvG-PointNet++ is by design vulnerable to PGD attacks only targeting on $\mathcal{L}_{gather}$ as such perturbations will potentially affect most gather vector predictions to make $\vg_i$ masked out so that the rest is insufficient for the final classification. 

\textbf{Experimentation.} We train GvG-PointNet++ based on single-scale grouped PointNet++~\cite{qi2017pointnet++} on ModelNet40 and set $r=0.08$ and $\lambda=1$ as suggested by~\cite{Dong_2020_CVPR}. The model is trained by Adam~\cite{kingma2014adam} optimizer with 250 epochs using batch size = 16, and the initial learning rate is 0.01. For the adaptive attack, we use 10-step binary search to find a appropriate $\beta$. The setup of $\normltwo$ norm-based PGD attacks is identical to~\cite{Dong_2020_CVPR}, and we also leverage $\normmax$ norm-based PGD-200 in the evaluation. Detailed setups are elaborated in Appendix A.1.








\textbf{Discussion.} As presented in Table~\ref{tb:adaptive_2}, both adaptive PGD attacks achieve high success rates on GvG-PointNet++. We also observe that the $\normmax$ norm-based PGD attack is more effective on $\mathcal{L}_{gather}$ since $\normmax$ norm perturbations assign the same adversarial budget to each point, which can easily influence a large number of gather vector predictions. However, it is hard for the $\normltwo$ norm-based PGD attack to affect so many gather vector predictions because it prefers to perturb key points (\ie points with larger gradients) rather than the whole point set. GvG-PointNet++ leverages DNN to detect adversarial perturbations, which is similar to MagNet~\cite{meng2017magnet} in the 2D space. We validate that adversarial detection also fails to provide true robustness under adaptive white-box adversaries in point cloud classification. 
\vspace{-0.2cm}
\section{AT with Different Symmetric Functions}
\label{sec:adv_train}

We have so far demonstrated that state-of-the-art defenses against 3D adversarial point clouds are still vulnerable to adaptive attacks. While gradient obfuscation cannot offer true adversarial robustness, adversarial training (AT) is widely believed as the most effective method. In this section, we conduct the first thorough study showing how AT performs in point cloud classification. 

\vspace{-0.2cm}
\subsection{Adversarial Training Preliminaries}
\label{sec:adv_train_pre}

Madry \etal \cite{madry2018towards} formulate AT as a paddle point problem in \Eqref{eq:adv_train}, where $\mathcal{D}$ is the underlying data distribution, $\mathcal{L}(\cdot,\cdot,\cdot)$ is the loss function, $\vx$ is the training data with its label $\vy$, $\vepsilon$ is the adversarial perturbation, and $\sS$ denotes the boundary of such perturbations. 
\begin{equation}
\argmin_{\vtheta} \quad \mathbb{E}_{(\vx,\vy)\sim{\mathcal{D}}}\left[ \max_{\vepsilon \in \sS} \mathcal{L}(\vx+\vepsilon, \vy, \vtheta) \right]
\label{eq:adv_train}
\end{equation}

\textbf{Adversarial training setups.} We choose PointNet~\cite{qi2017pointnet}\footnote{\url{https://github.com/charlesq34/pointnet}\label{pointnet}} as the primary backbone due to its extensive adoption in various 3D learning tasks~\cite{shi2019pointrcnn,lang2019pointpillars}. We also select DeepSets~\cite{NIPS2017_6931} and DSS~\cite{maron2020learning} to show the generality of our analysis. As illustrated in \Secref{sec:attack} and demonstrated by~\cite{madry2018towards}, $\normmax$ norm-based PGD attack (\Eqref{eq:pgd}) tends to be a universal first-order adversary. Thus, we select PGD-7 into the training recipe, and empirically set the maximum per-point perturbation $\epsilon=0.05$ out of the point cloud range $[-1,1]$. We follow the default PointNet training setting\textsuperscript{\ref{pointnet}} to train all models on the ModelNet40 training set. In the evaluation phase, we utilize PGD-200 to assess their robustness on the ModelNet40 validation set with the same adversarial budget $\epsilon=0.05$. Meanwhile, we also report the nominal accuracy on the clean validation set. Each PGD attack starts from a random point in the allowed perturbation space. More details can be found in Appendix B. 



\vspace{-0.1cm}
\subsection{AT with Fixed Pooling Operations}

As illustrated in \Figref{fig:spec}, point cloud classification models fundamentally follow a general specification $(\sigma \circ \rho \circ \Phi)(\sX)$. $\Phi(\cdot)$ represents a set of permutation-equivariant layers to learn local features from each point. $\rho(\cdot)$ is a column-wise symmetric (permutation-invariant) function to aggregate the learned local features into a global feature, and $\sigma(\cdot)$ are fully connected layers for final classification. PointNet, DeepSets, and DSS leverage different $\Phi(\cdot)$ for local feature learning, but all depend on \textbf{fixed pooling operations} as their $\rho(\cdot)$. Specifically, \verb+MAX+ pooling is by default used in DeepSets (for point cloud classification) and PointNet~\cite{NIPS2017_6931,qi2017pointnet}, while DSS utilizes \verb+SUM+ pooling~\cite{maron2020learning}. We also additionally select \verb+MEDIAN+ pooling due to its robust statistic feature~\cite{huber2004robust}. Though models with fixed pooling operations have achieved satisfactory accuracy under standard training, they face various difficulties under AT. As presented in Table~\ref{tb:fixed_adv_train}, models with \verb+MEDIAN+ pooling achieve better nominal accuracy among fixed pooling operations, but much worse adversarial accuracy, while \verb+SUM+ pooling performs contrarily. Most importantly, none of them reach a decent balance of nominal and adversarial accuracy.



\textbf{Analysis.} AT consists of two stages: 1) \textit{inner maximization} to find the worst adversarial examples and 2) \textit{outer minimization} to update model parameters. Fixed pooling operations essentially leverage a \textit{single} statistic to represent the distribution of a feature dimension~\cite{murray2014generalized}. Although \verb+MEDIAN+ pooling, as a robust statistic, intuitively should enhance the robustness, we find it actually hinders the inner maximization stage from making progress. We utilize $\normmax$ norm-based PGD attack to maximize the xent loss of standard trained model with three fixed pooling operations. \Figref{fig:loss} validates that \verb+MEDIAN+ pooling takes many more steps to maximize the loss. Therefore, \verb+MEDIAN+ pooling fails to find the worst adversarial examples in the first stage with limited steps. Though \verb+MAX+ and \verb+SUM+ pooling are able to achieve higher loss value, they encounter challenges in the second stage. \verb+MAX+ pooling backward propagates gradients to a \textit{single} point at each dimension so that the rest $\frac{n-1}{n}$ features do not contribute to model learning. Since $n$ is oftentimes a large number (\textit{e.g.,} 1024), the huge information loss and non-smoothness will fail AT~\cite{xie2020smooth}. While \verb+SUM+ pooling realizes a smoother backward propagation, it lacks discriminability because by applying the same weight to each element, the resulting representations are strongly biased by the adversarial perturbations. Thus, with \verb+SUM+ pooling, the models cannot generalize well on clean data (Table~\ref{tb:fixed_adv_train}).

\begin{figure}[t]
\vspace{-0.5cm}
\subfigure[Attention-based pooling operations apply self-attention to each point-level feature vector $\vf_i$. The learned weight $\alpha_i$ is multiplied with each element in $\vf_i$, and the aggregated feature is computed by a column-wise summation.\label{fig:att}]{\begin{minipage}[t]{0.53\linewidth}
\vspace{0.0cm}
\begin{center}
\includegraphics[width=\linewidth,height=2.5cm]{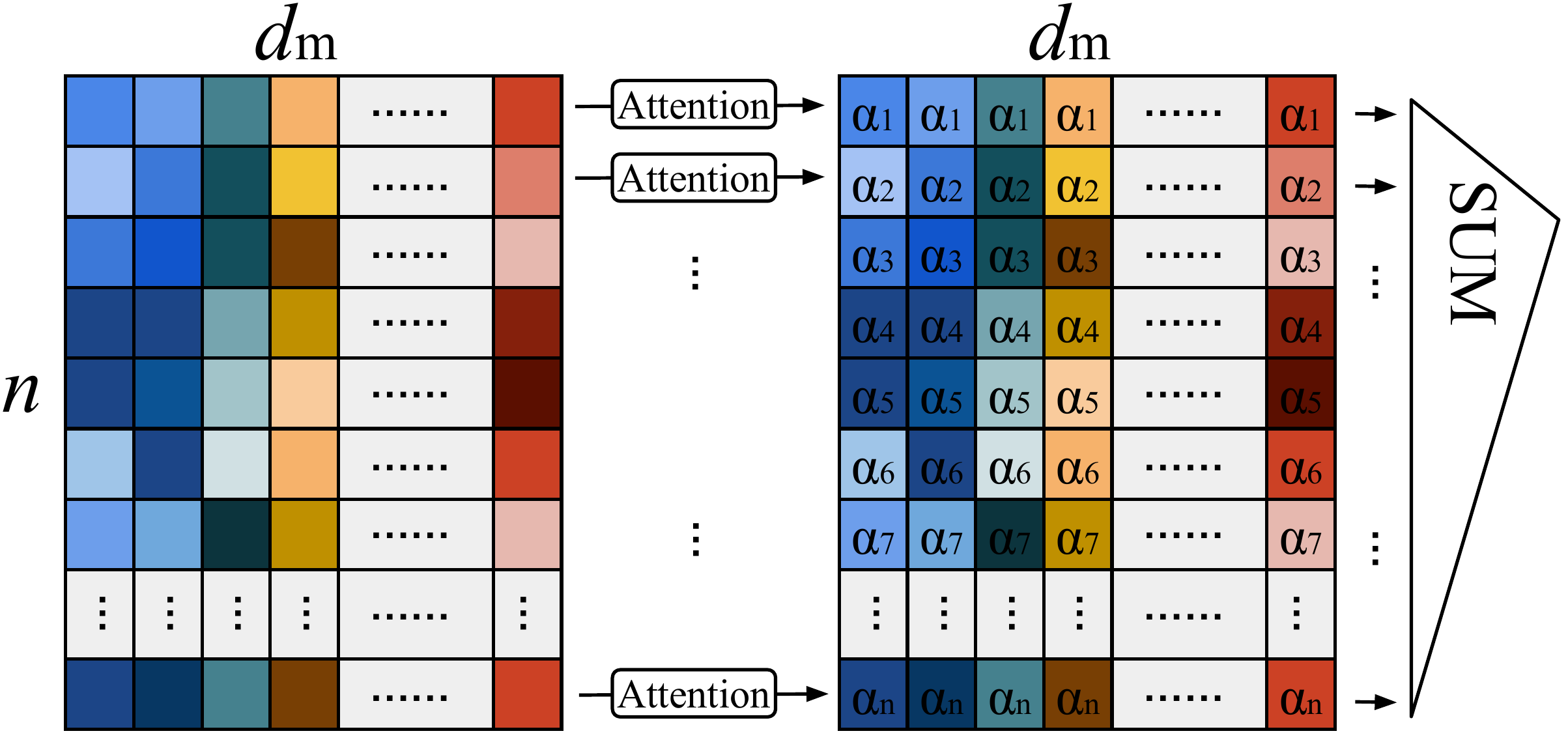} 
\end{center}
\end{minipage}}
\hfill
\subfigure[Sorting-based pooling operations sort each dimension to re-organize the feature set into an ordered matrix $\widetilde{\mF}$ to which complex operations (\textit{e.g.,} CNN) can be applied to aggregate features.\label{fig:sort}]{\begin{minipage}[t]{0.45\linewidth}
\vspace{-0.05cm}
\begin{center}
\includegraphics[width=\linewidth,height=2.5cm]{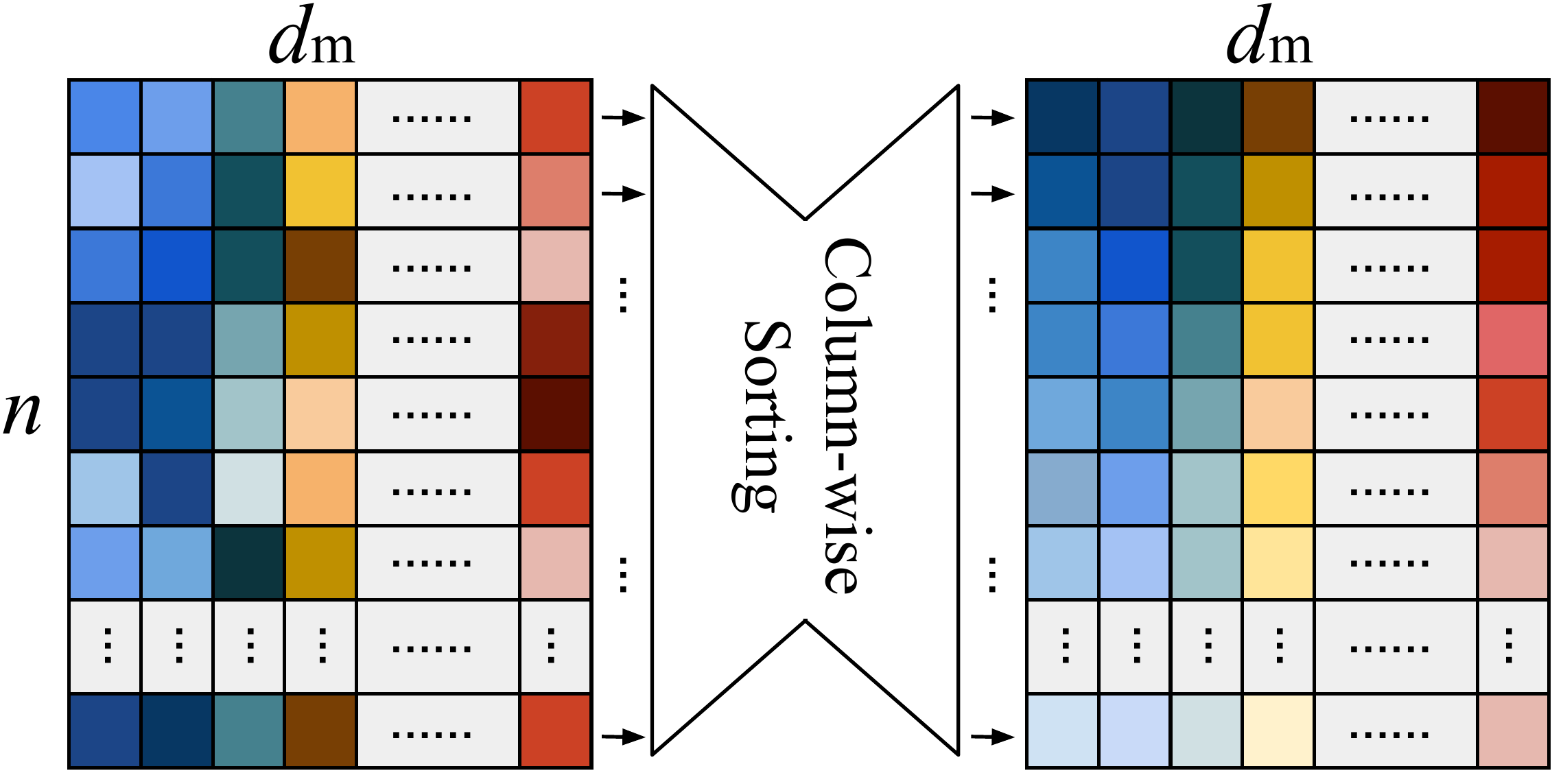} 
\end{center}
\end{minipage}}
\vspace{-0.3cm}
\caption{Design philosophy of attention-based and sorting-based pooling operations.}
\label{fig:att_sort}
\vspace{-0.5cm}
\end{figure}

\vspace{-0.1cm}
\subsection{AT with Parametric Pooling Operations}

Recent studies have also presented trainable \textbf{parametric pooling operations} for different tasks in set learning (\textit{e.g.,} multiple instance learning), which are also qualified to be the symmetric function $\rho(\cdot)$ in point cloud classification models. They can be further grouped into two classes: 1) \textit{attention-based} and 2) \textit{sorting-based} pooling. We apply them to our backbone models and further benchmark their robustness under AT. It is worth noting that \textit{none} of those parametric pooling operations are proposed to improve the adversarial robustness, and we are the first to conduct such an in-depth analysis of their performance as the symmetric function under AT in point cloud classification.

\vspace{-0.2cm}
\subsubsection{Attention-based Pooling Operations}
\label{sec:att}

An attention module can be abstracted as mapping a query and a set of key-value pairs to an output, making the models learn and focus on the critical information~\cite{bahdanau2014neural}. \Figref{fig:att} shows the design principle of attention-based pooling, which leverages a compatibility function to learn point-level importance. The aggregated global feature is computed as a column-wise weighted sum of the local features. Two attention-based pooling operations, \verb+ATT+ and \verb+ATT-GATE+, are first proposed for multiple instance learning~\cite{pmlr-v80-ilse18a}. Let $\sF = \{\vf_1, \vf_2, \dots, \vf_{n}\}$ be a set of features, \verb+ATT+ aggregates the global feature $\vg$ by:
\begin{equation}
    \vg = \sum_{i=1}^{n} a_i \cdot \vf_i\;,\quad a_i = \frac{\exp(\vw^\top \cdot \tanh(\mV \cdot {\vf}_i^\top))}{\sum_{j=1}^{n} \exp(\vw^\top \cdot \tanh(\mV \cdot {\vf}_j^\top))}
\end{equation}
where $\vw \in \R^{L \times 1}$ and $\mV \in \R^{L \times d_m}$ are learnable parameters. \verb+ATT-GATE+ improves the expressiveness of \verb+ATT+ by introducing another non-linear activation $\mathrm{sigmoid}(\cdot)$ and more trainable parameters into weight learning. Furthermore, \verb+PMA+~\cite{lee2019set} is proposed for general set learning, which leverages multi-head attention~\cite{vaswani2017attention} for pooling. We detail the design and our implementation of \verb+ATT+, \verb+ATT-GATE+, and \verb+PMA+ in Appendix B.3, and adversarially train the backbone models with these attention-based pooling operations.


\vspace{-0.2cm}
\subsubsection{Sorting-based Pooling Operations}
\label{sec:sorting}

Sorting has been recently considered in the set learning literature due to its permutation-invariant characteristic, as shown in~\Figref{fig:sort}. Let $\mF \in \R^{n \times d_m}$ be the matrix version of the feature set $\sF$, \verb+FSPool+~\cite{Zhang2020FSPool} aggregates $\mF$ by feature-wise sorting in a descending order:
\begin{equation}
    \widetilde{\mF}_{i,j} = {sort}_{\downarrow}({\mF_{:,j})}_i\;; \quad g_j = \sum_{i=1}^{n} \mW_{i,j} \cdot \widetilde{\mF}_{i,j}
\label{eq:fspool}
\end{equation}
where $\mW \in \R^{n \times d_m}$ are learnable parameters. Therefore, the pooled representation is column-wise weighted sum of $\widetilde{\mF}$. \verb+SoftPool+~\cite{wang2020softpoolnet} re-organizes $\mF$ so that its $j$-th dimension is sorted in a descending order, and picks the top $k$ point-level embeddings $\mF'_j \in \R^{k \times d_m}$ to further form $\widetilde{\mF}=[\mF'_1, \mF'_2, \dots, \mF'_{d_m}]$. Then, \verb+SoftPool+ applies CNN to each $\widetilde{\mF}_j \rightarrow \vg_j$ so that the pooled representation is $\vg = [\vg_1, \vg_2, \dots, \vg_{d_m}]$. Implementation details of \verb+SoftPool+ are elaborated in Appendix B.3 due to space limits. We also adversarially train the backbone models with \verb+FSPool+ and \verb+SoftPool+. 


\begin{figure*}[!t]
\begin{minipage}[t]{0.5\linewidth}
\renewcommand\arraystretch{1.2}
\setlength\tabcolsep{3pt}
\footnotesize
\vspace{0.2cm}
\captionof{table}{Adversarial robustness of models with parametric pooling operations under PGD-200 at $\epsilon=0.05$.}
\vspace{-0.2cm}
\label{tb:para_adv_train}
\begin{center}
\begin{tabular}{|c|c|c|c|c|c|c|}
\hline
\multirow{2}*{\makecell[c]{\bf Pooling \\ \bf Operation}} &\multicolumn{3}{c|}{\bf Nominal Accuracy} &\multicolumn{3}{c|}{\bf Adversarial Accuracy} \\
\cline{2-7}
&PointNet &DeepSets  &DSS  &PointNet &DeepSets  &DSS  \\
\hline
\hline
\verb+ATT+ &73.5\%   &52.3\%  &72.8\% &22.1\% &23.2\% &23.9\%\\
\hline
\verb+ATT-GATE+  &75.1\%   &63.9\%  &73.3\% &23.2\% &24.8\% &26.1\%\\
\hline
\verb+PMA+   &73.9\%   &51.9\% &72.5\% &25.4\% &20.9\% &23.9\%\\
\hline
\verb+FSPool+   &\textbf{82.8\%}   &73.8\% &81.5\% &29.8\% &25.3\% &26.1\%\\
\hline
\verb+SoftPool+   &79.8\%   &72.1\% &80.2\% &30.1\% &24.9\% &26.5\%\\
\hline
\makecell[c]{ \texttt{DeepSym} \\ (ours)}   &82.7\%   &\textbf{74.2\%} & \textbf{81.6\%} &\textbf{33.6\%} &\textbf{26.9\%} &\textbf{31.4\%}\\
\hline
\end{tabular}
\end{center}
\vspace{-0.4cm}

\end{minipage}
\hfill
\begin{minipage}[t]{0.47\linewidth}
\vfill
\renewcommand\arraystretch{1.1}
\setlength\tabcolsep{3pt}
\footnotesize
\captionof{table}{Adversarial robustness of PointNet with different pooling operations under attacks at $\epsilon=0.05$.}
\vspace{-0.4cm}
\label{tb:other_attack}
\begin{center}
\begin{tabular}{|c|c|c|c|c|c|c|}
\hline
\multirow{2}*{\makecell[c]{\bf Pooling \\ \bf Operation}} &\multicolumn{3}{c|}{\bf White-box Attack} &\multicolumn{3}{c|}{\bf Black-box Attack} \\
\cline{2-7}
&FGSM &BIM  &MIM  &SPSA &NES  &Evolution  \\
\hline
\hline
\verb+MAX+ &72.8\%   &24.3\%  &23.5\% &69.2\% &67.1\% &53.4\%\\
\hline
\verb+MEDIAN+ &\textbf{77.6\%}   &23.3\%  &14.5\% &71.1\% &65.2\% &57.8\%\\
\hline
\verb+SUM+ &44.4\%   &33.5\%  &37.5\% &65.3\% &62.3\% &52.7\%\\
\hline
\verb+ATT+ &43.1\%   &33.1\%  &35.0\% &68.1\% &64.8\% &55.9\%\\
\hline
\verb+ATT-GATE+  &43.9\%   &34.2\%  &33.9\% &70.2\% &65.9\% &55.8\%\\
\hline
\verb+PMA+   &47.2\%   &31.9\% &30.1\% &67.2\% &64.1\% &53.4\%\\
\hline
\verb+FSPool+   &61.3\%   &45.4\% &48.0\% &\textbf{72.8\%} &71.9\% &69.9\%\\
\hline
\verb+SoftPool+   &62.1\%   &47.6\% &45.1\% &69.2\% &68.5\% &70.0\%\\
\hline
\verb+DeepSym+ (ours)   &61.4\%   &\textbf{52.5\%} & \textbf{55.4\%} &72.4\% &\textbf{72.1\%} &\textbf{73.1\%}\\
\hline
\end{tabular}
\end{center}
\end{minipage}
\vspace{-0.5cm}
\end{figure*}

\vspace{-0.2cm}
\subsubsection{Experimental Analysis}
Table~\ref{tb:para_adv_train} presents the results of AT with different parametric pooling operations. To meet the requirement of permutation-invariance, attention-based pooling is restricted to learn \textit{point-level} importance. For example, \verb+ATT+ applies the same weight to all dimensions of a point embedding. As a result, attention barely improves the pooling operation's expressiveness as it essentially re-projects the point cloud to a single dimension (\textit{e.g.,} $\vf_i \rightarrow a_i$ in \verb+ATT+) and differentiates them based on it, which significantly limits their discriminability. Therefore, little useful information can be learned from the attention module, explaining why they perform similarly to \verb+SUM+ pooling that applies the same weight to each point under AT, as presented in Table~\ref{tb:para_adv_train}. Sorting-based pooling operations naturally maintain permutation-invariance as ${sort}_{\downarrow}(\cdot)$ re-organizes the unordered feature set $\sF$ to an ordered feature map $\widetilde{\mF}$. Thus, \verb+FSPool+ and \verb+SoftPool+ are able to further apply \textit{feature-wise} linear transformation and CNN. The insight is that feature dimensions are mostly independent of each other, and each point expresses to a different extent in every dimension. By employing feature-wise learnable parameters, the gradients also flow smoother through sorting-based pooling operations. Table~\ref{tb:para_adv_train} validates that sorting-based pooling operations achieve much better adversarial accuracy, \textit{e.g.,} on average, 7.3\% better than \verb+MAX+ pooling while maintaining comparable nominal accuracy. 

\vspace{-0.2cm}
\section{Improving Robustness with \texttt{DeepSym}}
\label{sec:deepsym}

In the above analysis, we have shed light on that sorting-based pooling operations can benefit AT in point cloud classification. We hereby explore to further improve the sorting-based pooling design based on existing arts. First, we notice that both \verb+FSPool+ and \verb+SoftPool+ apply ${sort}_{\downarrow}(\cdot)$ right after a ReLU function~\cite{nair2010rectified}. However, ReLU activation leads to half of neurons being zero~\cite{goodfellow2016deep}, which will 
make ${sort}_{\downarrow}(\cdot)$ unstable. Second, recent studies have shown that AT appreciates deeper neural networks~\cite{xie2019intriguing}. Nevertheless, \verb+FSPool+ only employs one linear layer to aggregate features, and \verb+SoftPool+ requires $d_m$ to be a small number. The reason is that scaling up the depth in these existing sorting-based pooling designs requires exponential growth of parameters, which will make the end-to-end learning intractable. 

To address the above limitations, we propose a simple yet powerful pooling operation, \verb+DeepSym+, that embraces the benefits of sorting-based pooling and also applies deep learnable layers to the pooling process. Given a feature set after ReLU activation $\sF \in {\R_+}^{n \times d_m}$, \verb+DeepSym+ first applies another linear transformation to re-map $\sF$ into $\R^{n \times d_m}$ so that $\vf'_i = \mW \cdot {\vf_i}^\top$ where $\mW \in \R^{d_m \times d_m}$ and $\sF' = \{\vf'_1, \vf'_2, \dots, \vf'_n\}$. Let $\mF'$ be the matrix version of $\sF'$, \verb+DeepSym+ 
further sorts $\mF'$ in a descending order (\Eqref{eq:fspool}) to get $\widetilde{\mF'}$. Afterwards, a column-wise shared MLP will be applied on the sorted feature map $\widetilde{\mF'}$: 
\begin{equation}
\begin{split}
     & g_j = \mathrm{MLP}(\widetilde{\mF'}_{:,j})\;, \quad j = \{1, 2, \dots, d_m \}
\end{split}
\label{eq:deepsym}
\end{equation}
to learn the global feature representation $\vg$. Each layer of the MLP composes of a linear transformation, a batch normalization module~\cite{ioffe2015batch}, and a ReLU activation function. Compared to \verb+FSPool+ that applies different linear transformations to different dimensions, \verb+DeepSym+ employs a \textit{shared} MLP to different dimensions. By doing so, \verb+DeepSym+ deepens the pooling process to be more capable of digesting the adversarial perturbations. \verb+DeepSym+ can also address the problem of \verb+SoftPool+ that is only achievable with limited $d_m$ because the MLP is shared by all the feature channels so that it can scale up to a large number of $d_m$ with little complexity increases (Table~\ref{tb:overhead}). Moreover, \verb+DeepSym+ is intrinsically flexible and general. For example, it clearly generalizes \verb+MAX+ and \verb+SUM+ pooling by specific weight vectors. Therefore, it can also theoretically achieve universality with $d_m \ge n$~\cite{wagstaff2019limitations} while being more expressive in its representation and smoother in gradients propagation. To deal with the variable-size point clouds, \verb+DeepSym+ adopts column-wise linear interpolation in $\widetilde{\mF'}$ to form a continuous feature map and then re-samples it to be compatible with the trained MLP~\cite{jaderberg2015spatial}. Last but not least, \verb+DeepSym+ is by design flexible with its own number of layers and number of pooled features from each dimension. It is easy to tune the depth and width of \verb+DeepSym+ across different learning tasks. In the paper, we only allow DeepSym to output a single feature for a fair comparison with others. However, it is hard for other pooling operations to achieve this ability. For example, it requires a linear complexity increase for \verb+FSPool+ to enable this capability.


\begin{figure*}[t]
\vspace{-0.5cm}
\subfigure[PGD-1 adversarial training on ModelNet10.\label{fig:deepsym_1}]{\begin{minipage}[t]{0.43\linewidth}
\begin{center}
\includegraphics[width=\linewidth]{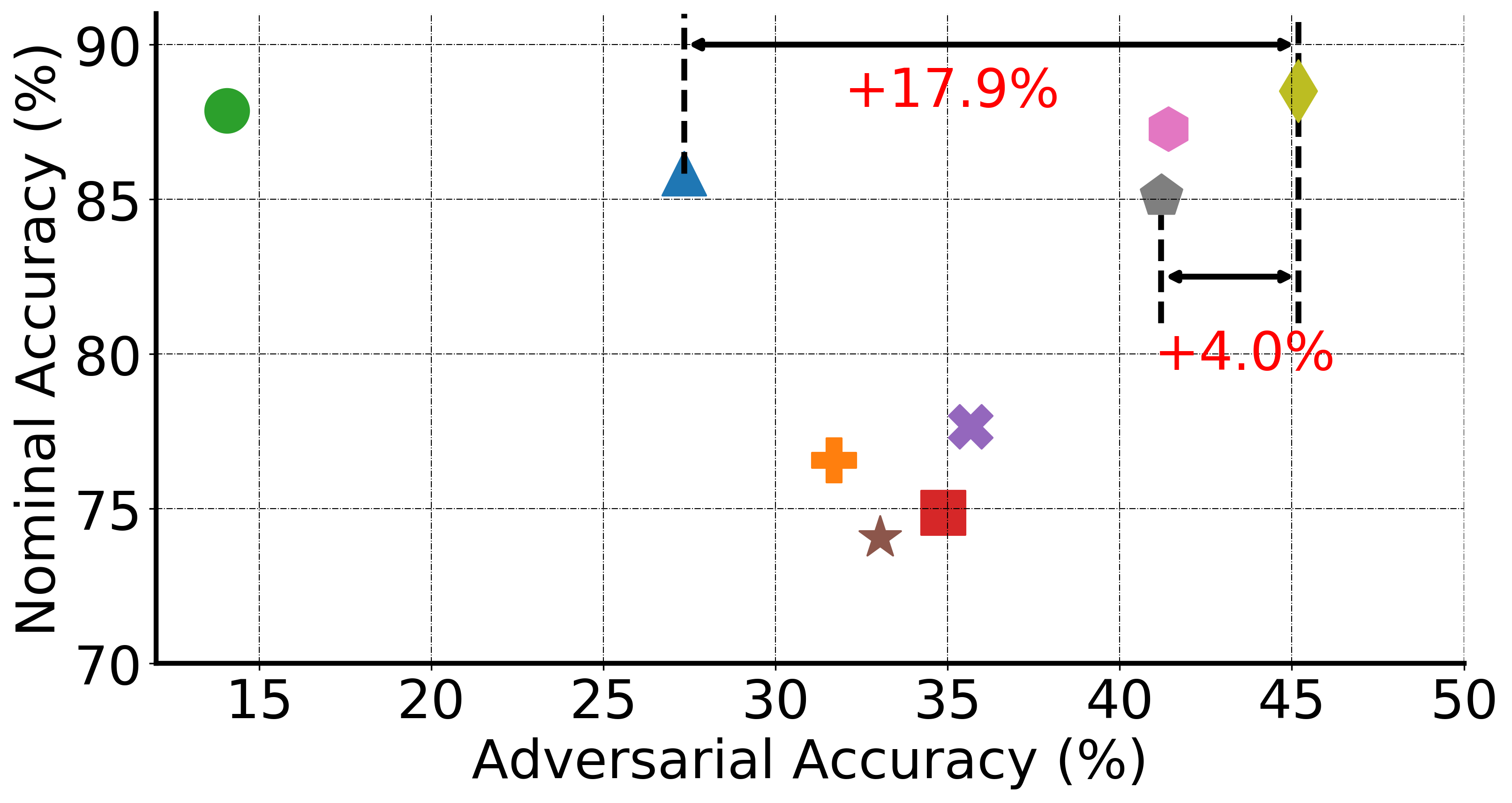} 
\end{center}
\end{minipage}}
\hfill
\subfigure[PGD-20 adversarial training on ModelNet40.\label{fig:deepsym_2}]{\begin{minipage}[t]{0.56\linewidth}
\begin{center}
\includegraphics[width=\linewidth]{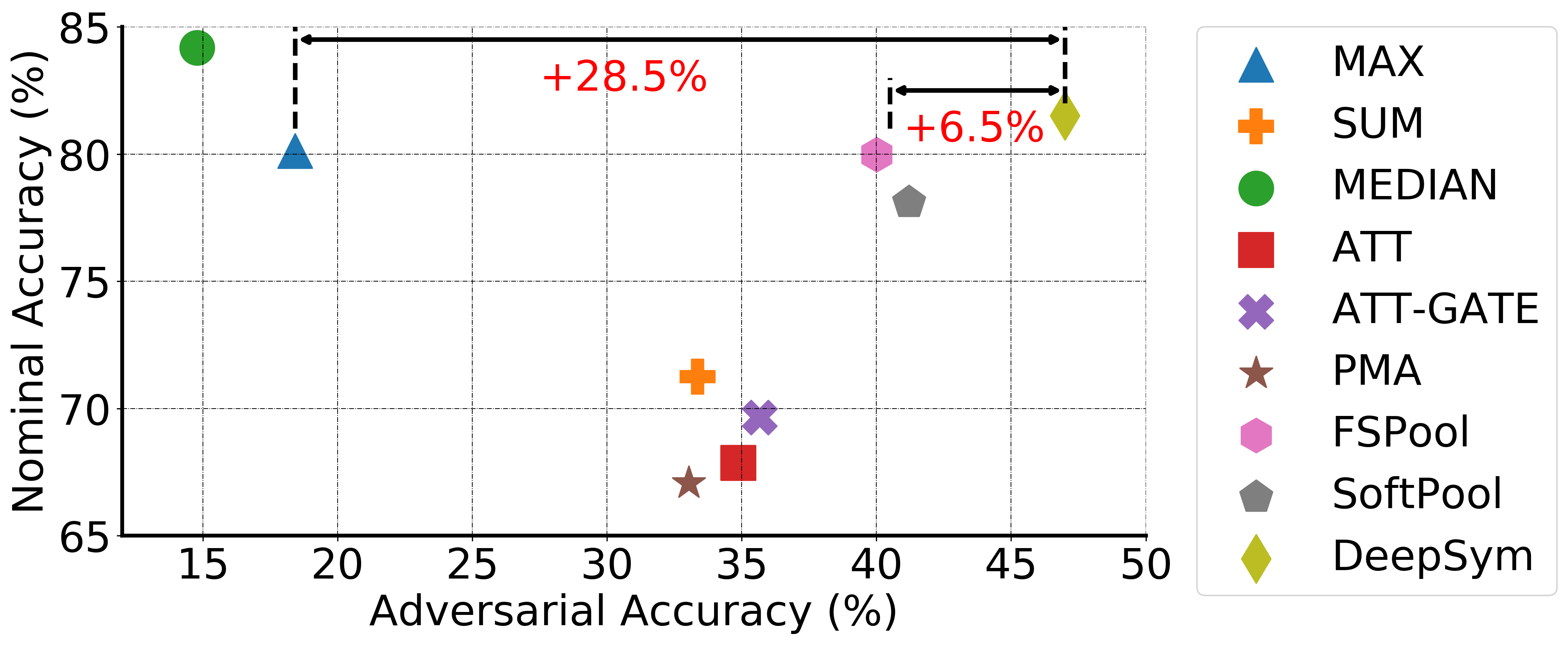} 
\end{center}
\end{minipage}}
\vspace{-0.2cm}
\caption{Adversarial robustness of PointNet with various pooling operations under PGD-200 at $\epsilon=0.05$.}
\label{fig:deepsym}
\vspace{-0.4cm}
\end{figure*}

\vspace{-0.1cm}
\subsection{Evaluations}
\vspace{-0.1cm}

As \verb+DeepSym+ is naturally flexible, we implement a 5-layer \verb+DeepSym+ with $[512,128,32,8,1]$ hidden neurons on three backbone networks and adversarially train them on ModelNet40 the same way introduced in \Secref{sec:adv_train_pre}. Table~\ref{tb:para_adv_train} shows that almost all models with \verb+DeepSym+ reach the best results in both nominal and adversarial accuracy, outperforming the default architecture by 10.8\%, on average. Taking PointNet as an example, \verb+DeepSym+ (33.6\%) improves the adversarial accuracy by 17.5\% ($\sim 2.1\times$) compared to the original \verb+MAX+ pooling architecture. Besides, \verb+DeepSym+ also achieves a 3.5\% improvement in adversarial accuracy compared to \verb+FSPool+ and \verb+SoftPool+. Overall, we demonstrate that \verb+DeepSym+ can benefit AT significantly in point cloud classification. 

We further leverage various white- and black-box adversarial attacks to cross validate the robustness improvements of \verb+DeepSym+ on PointNet. Specifically, we exploit well-known FGSM~\cite{szegedy2013intriguing}, BIM~\cite{kurakin2016adversarial}, and MIM~\cite{dong2018boosting} as the white-box attack methods. We set the adversarial budget $\epsilon = 0.05$, and leverage 200 steps for the iterative attacks, as well. For the black-box attacks, we choose two score-based methods: SPSA~\cite{uesato2018adversarial} and NES~\cite{ilyas2018black}, and a decision-based evolution attack~\cite{dong2020benchmarking}. We still select $\epsilon = 0.05$ and allow 2000 queries to find each adversarial example. The detailed setups are elaborated in Appendix C.1. As shown in Table~\ref{tb:other_attack}, PointNet with \verb+DeepSym+ consistently achieves the best adversarial accuracy under white-box attacks, except for FGSM. The reason is that FGSM is a single-step method that has limited ability to find adversarial examples (\Figref{fig:loss}). Besides, we find the black-box attacks are not as effective as the white-box attacks, which also demonstrate that adversarial training with \verb+DeepSym+ is able to improve the robustness of point cloud classification without gradient obfuscation~\cite{carlini2019evaluating}.


\begin{table}[!h]
\renewcommand\arraystretch{1.1}
\setlength\tabcolsep{4pt}
\footnotesize
\vspace{-0.35cm}
\caption{Overhead measurement of PointNet with different pooling operations.}
\vspace{-0.45cm}
\label{tb:overhead}
\begin{center}
\begin{tabular}{|c|c|c|c|}
\hline
\makecell[c]{\bf Pooling \\ \bf Operation} &\makecell[c]{\bf Inference Time \\ \bf (ms)} & \makecell[c]{\bf \# Trainable \\ \bf Parameters}  &\makecell[c]{\bf GPU Memory \\ \bf (MB)}  \\
\hline
\hline
\verb+MAX+ &2.21   &815,336  &989 \\
\hline
\verb+MEDIAN+ &2.44   &815,336  &989 \\
\hline
\verb+SUM+ &2.23  &815,336  &989 \\
\hline
\verb+ATT+ &2.71   &1,340,649  &1980 \\
\hline
\verb+ATT-GATE+  &3.07   &1,865,962  &2013 \\
\hline
\verb+PMA+   &\textbf{2.10}   &652,136 &981 \\
\hline
\verb+FSPool+   &2.89   &1,863,912 &1005 \\
\hline
\verb+SoftPool+   &2.85   &\textbf{355,328} &\textbf{725} \\
\hline
\verb+DeepSym+ (ours)   &3.10   &1,411,563 &2013 \\
\hline
\end{tabular}
\end{center}
\vspace{-0.5cm}
\end{table}

Since \verb+DeepSym+ brings deep trainable layers into the original backbones, it is necessary to report its overhead. We leverage TensorFlow~\cite{abadi2016tensorflow} and NVIDIA profiler~\cite{nvidia} to measure the inference time, the number of trainable parameters, and GPU memory usage on PointNet. Specifically, the inference time is averaged from 2468 objects in the validation set, and the GPU memory is measured on an RTX 2080 with batch size = 8. As shown in Table~\ref{tb:overhead}, \verb+DeepSym+ indeed introduces more computation overhead by leveraging the shared MLP. However, we believe the overhead is relatively small and acceptable, compared to its massive improvements on the adversarial robustness. To further have a lateral comparison, point cloud classification backbones are much more light-weight than image classification models. For example, ResNet-50~\cite{he2016deep} and VGG-16~\cite{simonyan2014very} have 23 and 138 million trainable parameters, respectively, and take much longer time to do the inference. The reason that models with \verb+SoftPool+ and \verb+PMA+ have fewer trainable parameters is that they limit the number of dimensions in the global feature by design.

\vspace{-0.1cm}
\subsection{Exploring The Limits of \textbf{\texttt{DeepSym}}}

There is a trade-off between the training cost and adversarial robustness in AT. Increasing the number of PGD attack steps can create harder adversarial examples~\cite{madry2018towards}, which could further improve the model's robustness. However, the training time also increases linearly with the number of attack iterations increasing. Due to PointNet's broad adoption~\cite{guo2020deep}, we here analyze how it performs under various AT settings. Specifically, we exploit the most efficient AT with PGD-1 on ModelNet10~\cite{Wu_2015_CVPR}, a dataset consisting of 10 categories with 4899 objects, and a relatively expensive AT with PGD-20 on ModelNet40 to demonstrate the effectiveness of \verb+DeepSym+. Other training setups are identical to \Secref{sec:adv_train_pre}. We also find the PointNet implementation\textsuperscript{\ref{pointnet}} default leverages random rotation for data augmentation to improve its isometric stability. As we do not take isometric robustness into consideration, we further remove such augmentation to simplify the learning task. We then leverage the default setting (PGD-7) to adversarially train the model and test its robustness.

\begin{figure}[h]
\vspace{-0.2cm}
\begin{center}
\includegraphics[width=\linewidth]{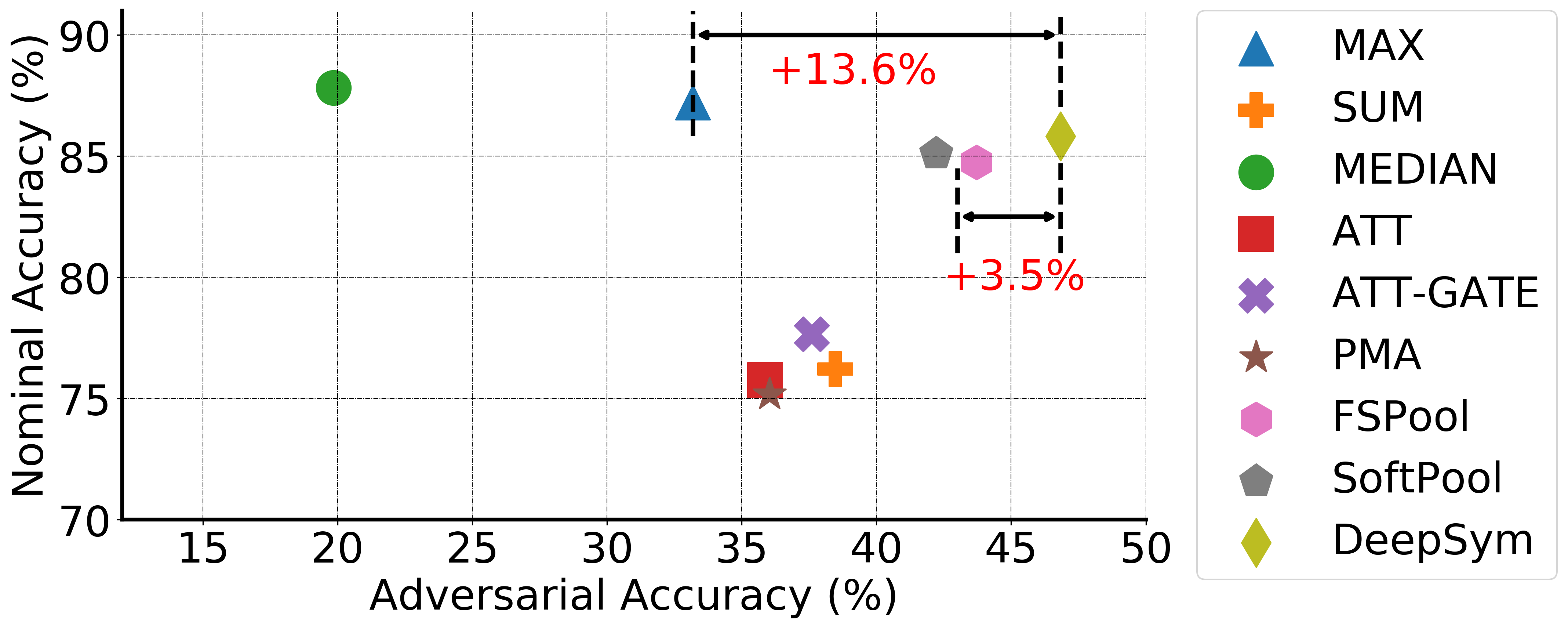} 
\end{center}
\vspace{-0.5cm}
\caption{Robustness of PointNet on ModelNet40 without rotation augmentation under PGD-200 at $\epsilon=0.05$.}
\label{fig:no_rot}
\end{figure}

\Figref{fig:deepsym} shows the results of the robustness of adversarially trained PointNet with various pooling operations under PGD-200. We demonstrate that PointNet with \verb+DeepSym+ still reaches the best adversarial accuracy of 45.2\% under AT with PGD-1 on ModelNet10, which outperforms the original \verb+MAX+ pooling by 17.9\% ($\sim 1.7\times$) and \verb+SoftPool+ by 4.0\%. Surprisingly, PointNet with \verb+DeepSym+ also achieves the best nominal accuracy of 88.5\%. Moreover, \verb+DeepSym+ further advances itself under AT with PGD-20 on ModelNet40. \Figref{fig:deepsym_2} shows that PointNet with \verb+DeepSym+ reaches the best 47.0\% adversarial accuracy, which are 28.5\% ($\sim 2.6\times$) and 6.5\% improvements compared to \verb+MAX+ pooling and \verb+SoftPool+, respectively while maintaining competent nominal accuracy. \Figref{fig:no_rot} shows that the baseline AT results improve due to the simplicity without rotation, as expected. Nevertheless, \verb+DeepSym+ still outperforms other pooling operations by a significant margin (13.6\%). We also report detailed evaluations using different PGD attack steps and budgets $\epsilon$ in Appendix C.1.

\vspace{-0.2cm}
\section{Concluding Remarks}
\label{sec:conclusion}
\vspace{-0.2cm}
In this work, we perform the \textit{first} rigorous study on the adversarial robustness of point cloud classification. We design adaptive attacks and demonstrate that state-of-the-art defenses cannot provide true robustness. Furthermore, we conduct a thorough analysis of how the required symmetric function affects the AT performance of point cloud classification models. We are the first to identify that the fixed pooling generally weakens the models' robustness under AT, and on the other hand, \textit{sorting-based} parametric pooling benefits AT well. Lastly, we propose \verb+DeepSym+ that further architecturally advances the adversarial accuracy of PointNet to 47.0\% under AT, outperforming the original design and a strong baseline by 28.5\% ($\sim 2.6 \times$) and 6.5\%. 

{\small
\bibliographystyle{ieee_fullname}
\bibliography{iclr2021_conference}
}
\appendix
\section{Adaptive Attack Experimental Setups and Visualizations}

\subsection{Experimental Setups}
\label{ap:adap_setup}
Since \textbf{DUP-Net}~\cite{Zhou_2019_ICCV} is open-sourced, we target the publicly released PointNet and PU-Net models. For the $\normltwo$ norm-based C\&W attack~\cite{Carlini_2017}, we set the loss function as:
\begin{equation}
    \mathcal{L} = (\max_{i \neq t'}(\mathcal{Z}(\mX')_{i}) - \mathcal{Z}(\mX')_{t'})^+ + \lambda \cdot || \mX - \mX' ||_2
\end{equation}
where $\mX \in \R^{n \times 3}$ is the matrix version of point cloud $\sX$, $\mX'$ is the optimized adversarial example, $\mathcal{Z}(\mX)_i$ is the $i$-th element of the output logits, and $t'$ is the target class. We leverage 10-step binary search to find the appropriate hyper-parameter $\lambda$ from $[10,80]$. As suggested by~\cite{xiang2019generating}, we choose 10 distinct classes and pick 25 objects in each class from the ModelNet40 validation set for evaluation. The step size of the adversarial optimization is 0.01 and we allow at most 500 iterations of optimization in each binary search to find the adversarial examples.

For the $\normmax$ norm-based PGD attack, we adopt the formulation in~\cite{madry2018towards}:  
\begin{equation}
    \mX_{t+1} = \Pi_{\mX + \mathcal{S}}(\mX_{t} + \alpha \cdot \sign(\nabla_{\mX_{t}} \mathcal{L}(\mX_t,\vtheta,\vy)))
    \label{eq:pgd}
\end{equation}
where $\mX_t$ is the adversarial example in the $t$-th attack iteration, $\Pi$ is the projection function to project the adversarial example to the pre-defined perturbation space $\mathcal{S}$, which is the $\normmax$ norm ball in our setup, and $\alpha$ is the step size. The $\text{sign}()$ function also normalizes the perturbation into the $\normmax$ norm ball for each iteration.  We select the boundary of allowed perturbations $\epsilon = \{0.01, 0.025, 0.05, 0.075\}$ out of the point cloud data range $[-1,1]$. Since point cloud data is continuous, we set the step size $\alpha = \frac{\epsilon}{10}$. 

For \textbf{GvG-PointNet++}~\cite{Dong_2020_CVPR}, we train it based on the single scale grouping (SSG)-PointNet++ backbone. The backbone network has three PointNet set abstraction module to hierarchically aggregate local features, and we enable gather vectors in the last module, which contains 128 local features (\textit{i.e.,} $n' = 128$ in Section 3.2) with 256 dimensions. To learn the gather vectors, we apply three fully connected layers with 640, 640, and 3 hidden neurons respectively, as suggested by Dong~\etal~\cite{Dong_2020_CVPR}. Since the data from ModelNet40 is normalized to [-1,1], the global object center is $\vc_g = [0,0,0]$. 

For the $\normmax$ norm-based PGD attack, we leverage the same setup as the attack on DUP-Net. For the $\normltwo$ norm-based PGD attack, we follow the settings in~\cite{Dong_2020_CVPR} to set the $\normltwo$ norm threshold $\epsilon = \delta \sqrt{n \times d_{in}}$, where $\delta$ is selected in $\{0.08, 0.16, 0.32\}$, $n$ is the number of points, and $d_{in}$ is the dimension of input point cloud (\textit{i.e.,} 3). The attack iteration is set to 50, and the step size $\alpha = \frac{\epsilon}{50}$.

\subsection{Visualizations}
\label{ap:adap_vis}

\begin{figure*}[h]
\begin{center}
\includegraphics[width=\linewidth]{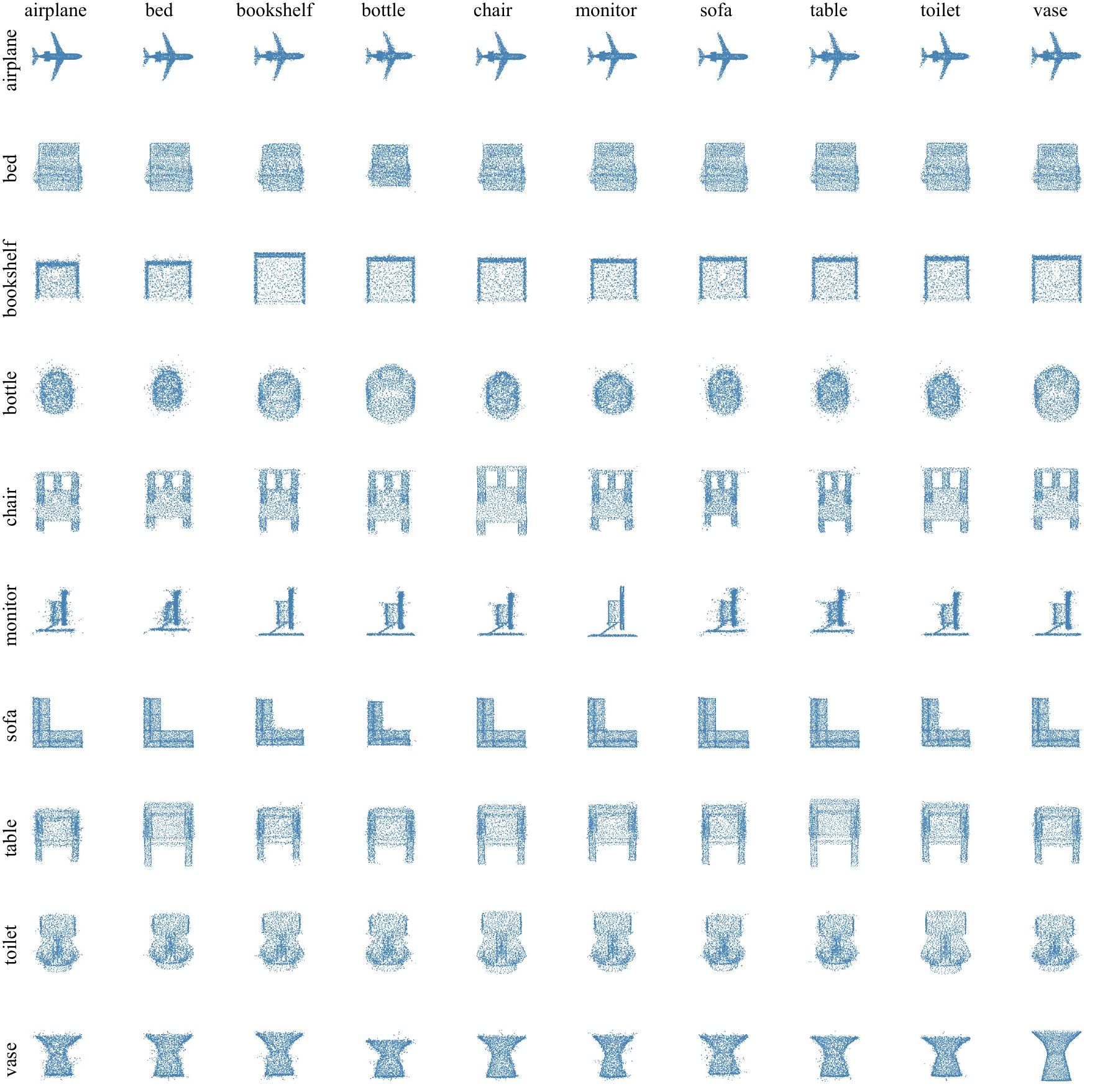} 
\end{center}
\caption{Visualizations of adversarial examples (2048 points) generated by $\normltwo$ norm-based C\&W adaptive attacks on \underline{PU-Net}.}
\label{fig:vis_pu}
\vspace{-0.5cm}
\end{figure*}

\begin{figure*}[h]
\begin{center}
\includegraphics[width=\linewidth]{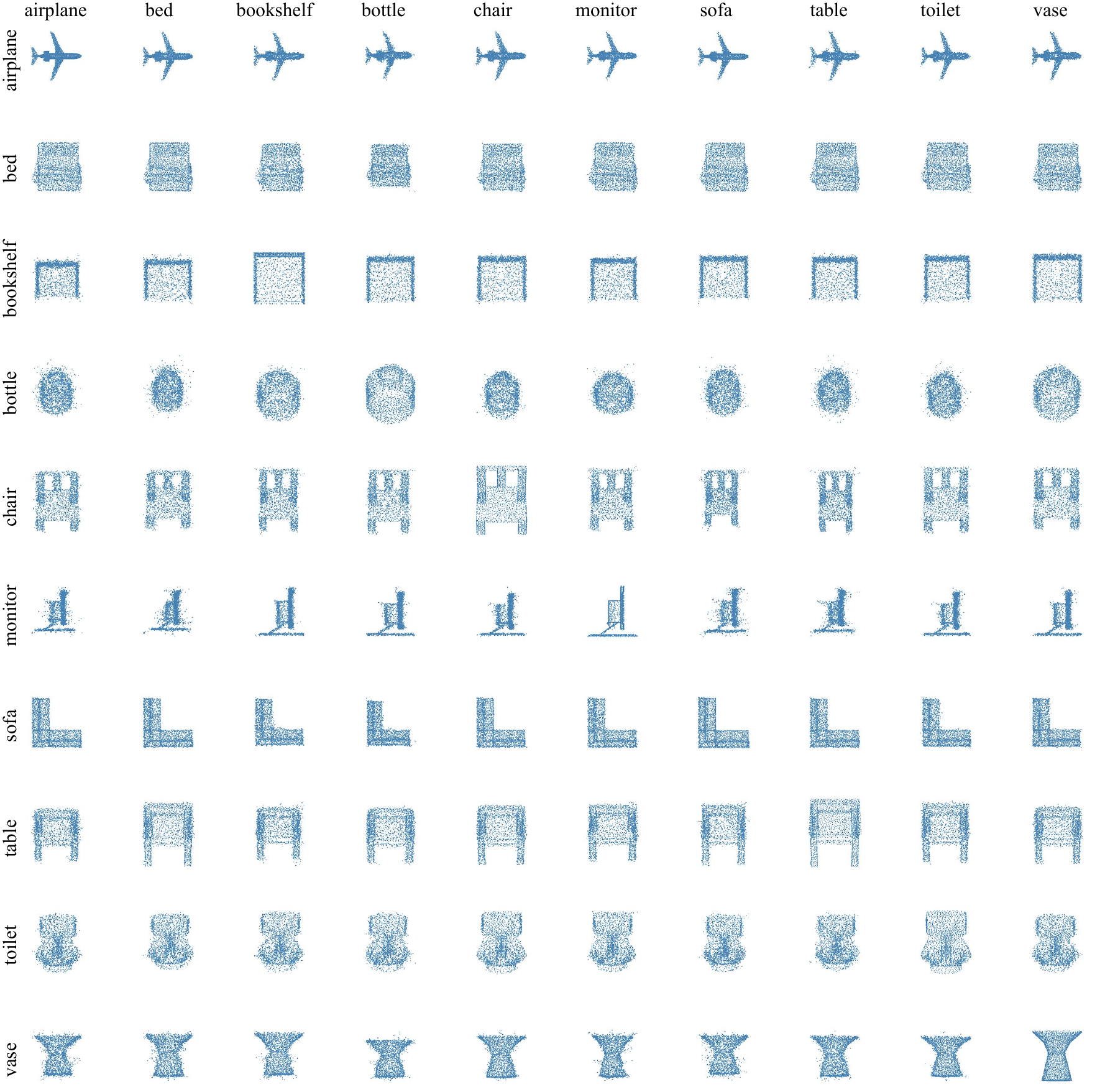} 
\end{center}
\caption{Visualizations of adversarial examples (2048 points) generated by $\normltwo$ norm-based C\&W adaptive attacks on \underline{DUP-Net}.}
\label{fig:vis_dup}
\end{figure*}

\begin{figure}[h]
\begin{center}
\includegraphics[width=\linewidth]{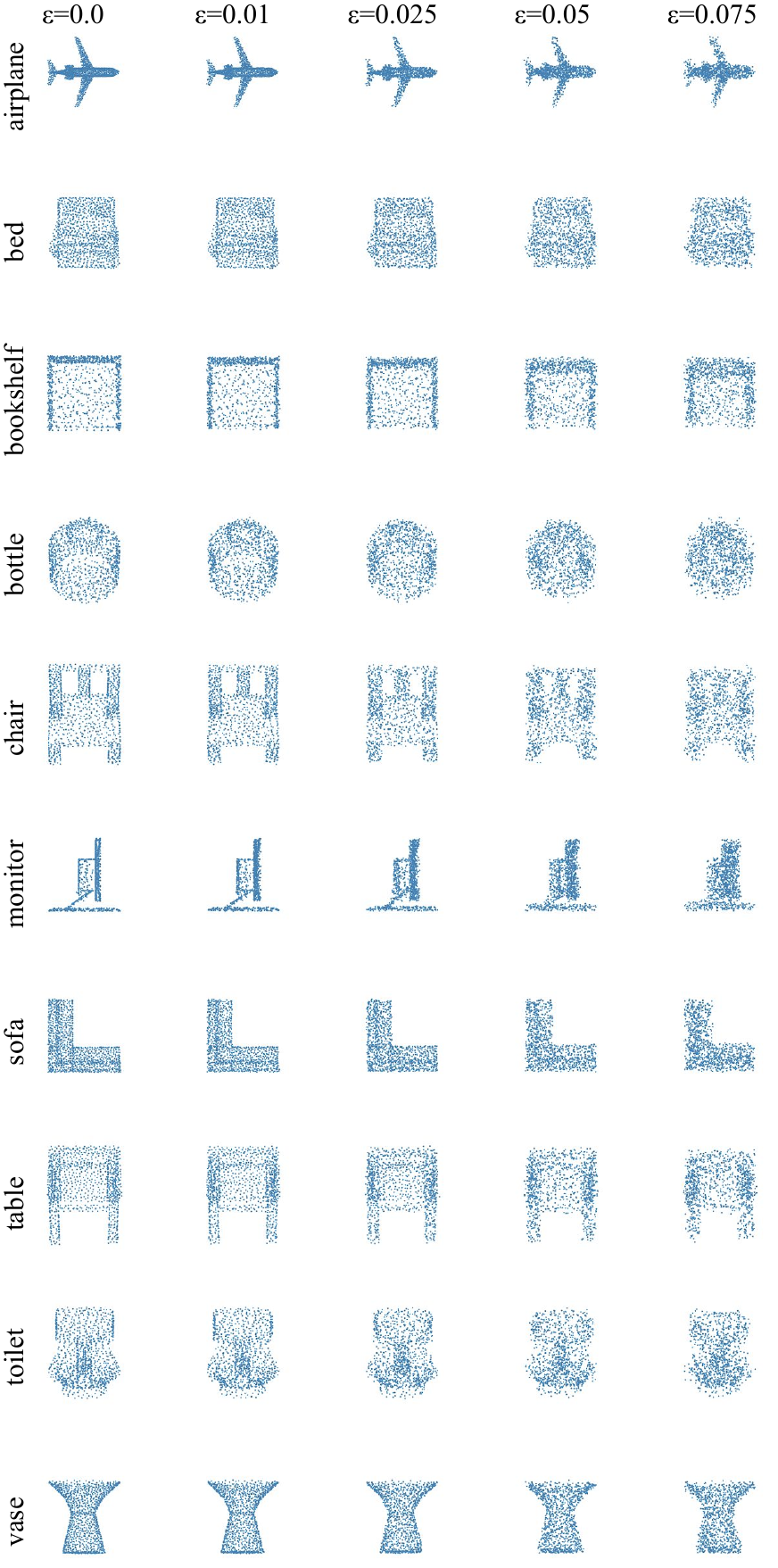} 
\end{center}
\caption{Visualizations of adversarial examples (1024 points) generated by $\normmax$ norm-based PGD adaptive attacks on \underline{GvG-PointNet++}.}
\label{fig:vis_gvg}
\end{figure}

We visualize some adversarial examples generated by adaptive attacks on PU-Net and DUP-Net in \Figref{fig:vis_pu} and \Figref{fig:vis_dup}. It is expected that adversarial examples targeting DUP-Net are noisier than the ones targeting PU-Net as the former needs to break the denoiser layer. However, as mentioned in Section 3.1, they are barely distinguishable from human perception. We also visualize some adversarial examples generated by untargeted adaptive PGD attacks on GvG-PointNet++ in \Figref{fig:vis_gvg} with different perturbation budgets $\epsilon$.  

\section{Adversarial Training Setup}
\label{ap:at}

\subsection{PGD Attack in Adversarial Training}

We also follow the formulation in \Eqref{eq:pgd} to find the worst adversarial examples. Specifically, we empirically select $\epsilon = 0.05$ into the training recipe as there is no quantitative study on how much humans can bear the point cloud perturbations.
\Figref{fig:vis_gvg} shows that adversarial examples with $\epsilon = 0.05$ are still recognizable by human perception. Moreover, because point cloud data is continuous, we set the step size of PGD attacks as:
\begin{equation}
    \alpha = \left\{
    \begin{aligned}
        & \frac{\epsilon}{\mathrm{step}},\quad \mathrm{step} < 10   \\
        & \frac{\epsilon}{10}, \quad \mathrm{step} \geq 10
    \end{aligned}
    \right.
\end{equation}
in both training and evaluation phases to make sure PGD attacks reach the allowed maximum perturbations.

\subsection{Point Cloud Classification Model Architecture Details}

PointNet~\cite{qi2017pointnet}, DeepSets~\cite{NIPS2017_6931}, and DSS~\cite{maron2020learning} are the fundamental architectures in point cloud classification. Other models, such as PointNet++~\cite{qi2017pointnet++} and DGCNN~\cite{dgcnn}, are built upon PointNet and DeepSets. Moreover, complex models oftentimes apply non-differentiable layers like $knn(\cdot)$ into end-to-end learning, which will make the adversarial training ineffective. In this work, we aim at exploring how the symmetric (permutation-invariant) function can benefit adversarial training. To this end, we choose PointNet, DeepSets, and DSS as the backbone networks. For the ModelNet40 dataset, we follow the default setting to split into 9,843 objects for training and 2,468 objects for validation~\cite{Wu_2015_CVPR}. We randomly sample 1024 points from each object to form its point cloud, if not otherwise stated. 

\textbf{PointNet.} We leverage the default architecture in PointNet codebase\footnote{\url{https://github.com/charlesq34/pointnet}} and exclude the transformation nets (\textit{i.e.,} T-Net) and dropout layers for simplicity and reproducibility. PointNet leverages shared fully connected (FC) layers as the permutation-equivariant layer $\phi_l: \mathrm{FC}_l({\mF_l}_{:,i}) \rightarrow {\mF_{l+1}}_{:,i}$ and \verb+MAX+ pooling as the symmetric function $\rho(\cdot)$. 

\textbf{DeepSets.} We leverage the default architecture in DeepSets codebase\footnote{\url{https://github.com/manzilzaheer/DeepSets}}. Different from PointNet, DeepSets first applies a symmetric function to each feature map and aggregate it with the original feature map. Afterwards, DeepSets also leverages FC layers to further process the features: $\phi_{l}: \mathrm{FC}_l({\mF_l}_{:,i} - \zeta(\mF_l)) \rightarrow {\mF_{l+1}}_{:,i}$, where $\zeta(\cdot)$ is column-wise \verb+MAX+ pooling in the original implementation. Similarly, \verb+MAX+ pooling is still used as $\rho(\cdot)$ in DeepSets.

\textbf{DSS.} DSS generalizes DeepSets architecture and applies another FC layer to $\zeta(\mF_l)$ in DeepSets so that $\phi_{l}: {\mathrm{FC}_l}_1({\mF_l}_{:,i}) + {\mathrm{FC}_l}_2(\zeta(\mF_l)) \rightarrow {\mF_{l+1}}_{:,i}$. Different from other two achitectures, DSS utilizes \verb+SUM+ pooling as $\rho(\cdot)$. Since there is no available codebase at the time of writing, we implement DSS by ourselves.

\begin{figure*}[!tp]
\centering
\subfigure[$\phi(\cdot)$ in PointNet.]{
\begin{minipage}[t]{0.4\linewidth}
\centering
\includegraphics[width=\linewidth]{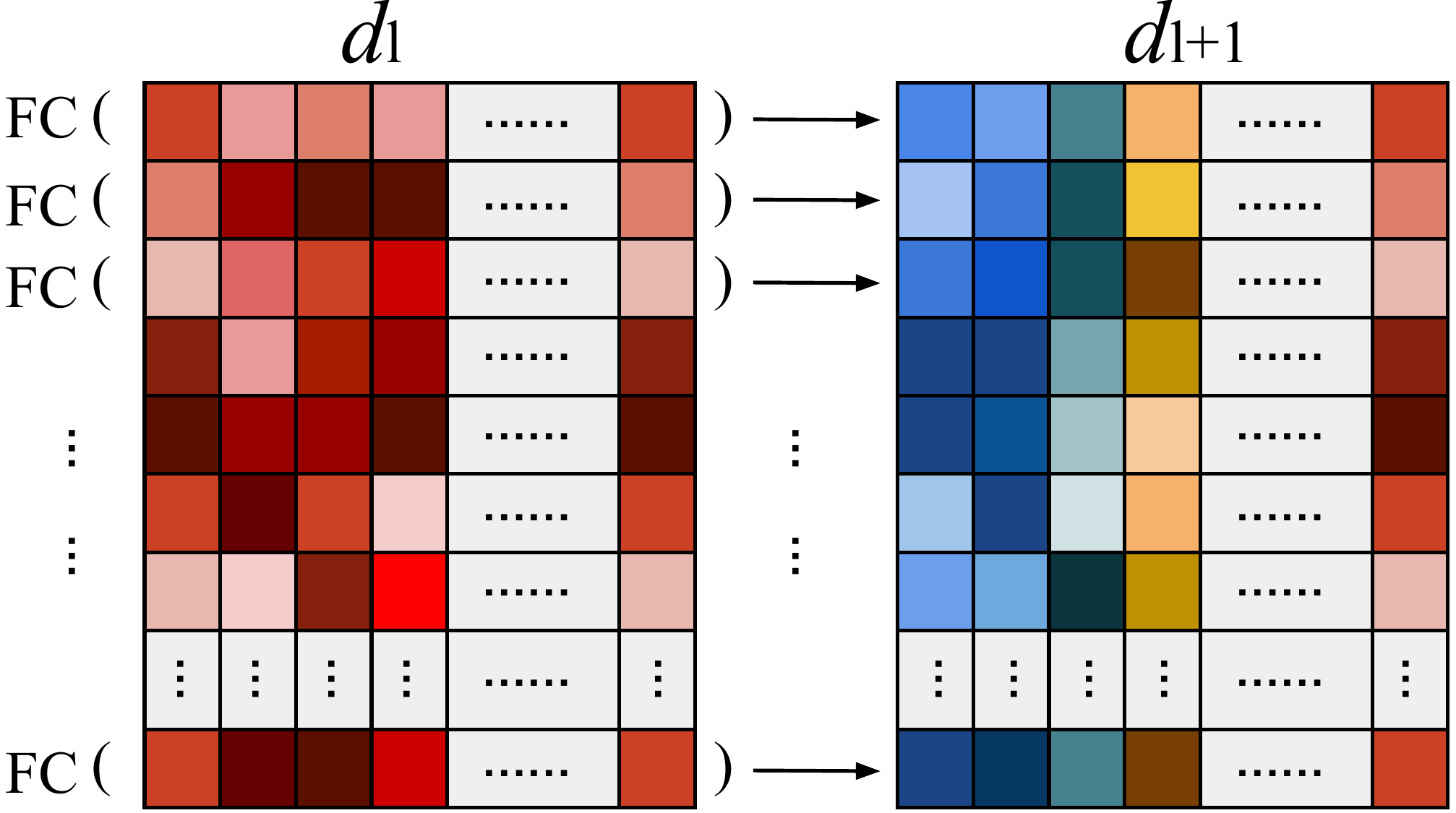}
\end{minipage}
}%
\subfigure[$\phi(\cdot)$ in DeepSets.]{
\begin{minipage}[t]{0.56\linewidth}
\centering
\includegraphics[width=\linewidth]{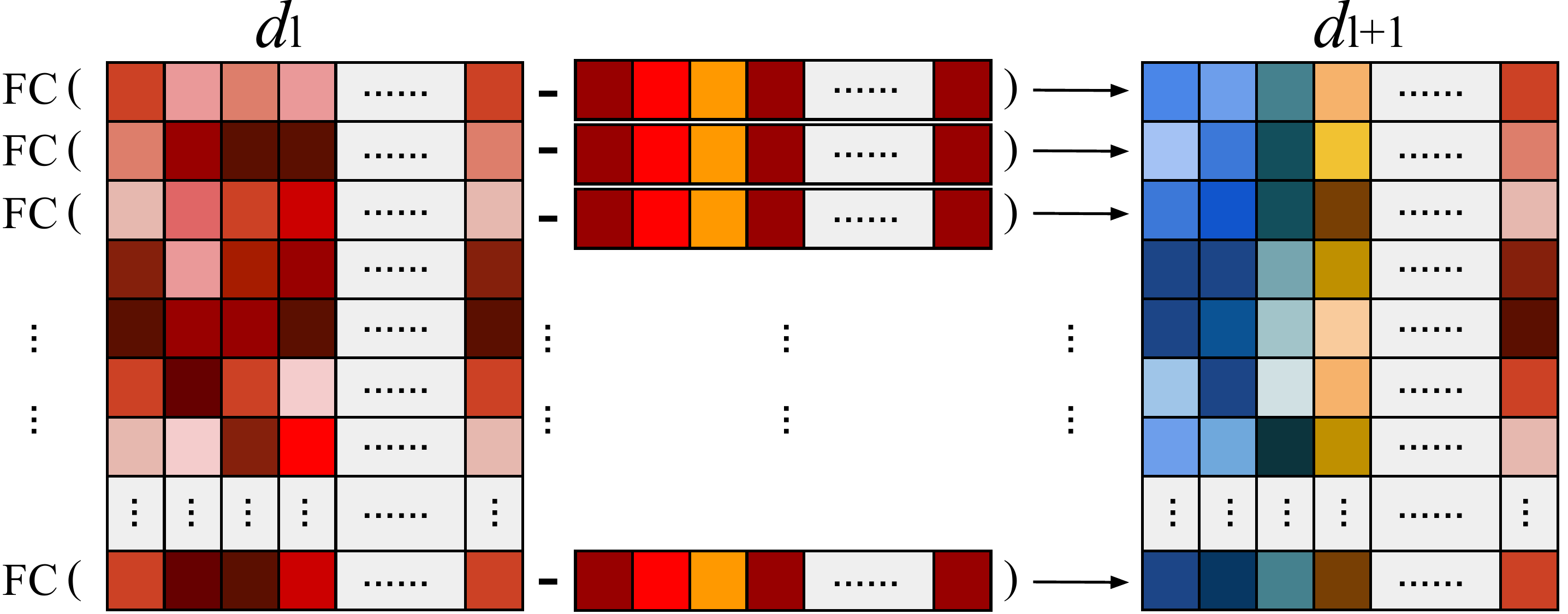}
\end{minipage}%
}%
\vskip 0pt
\vspace{-0.3cm}
\subfigure[$\phi(\cdot)$ in DSS.]{
\begin{minipage}[t]{0.58\linewidth}
\centering
\includegraphics[width=\linewidth]{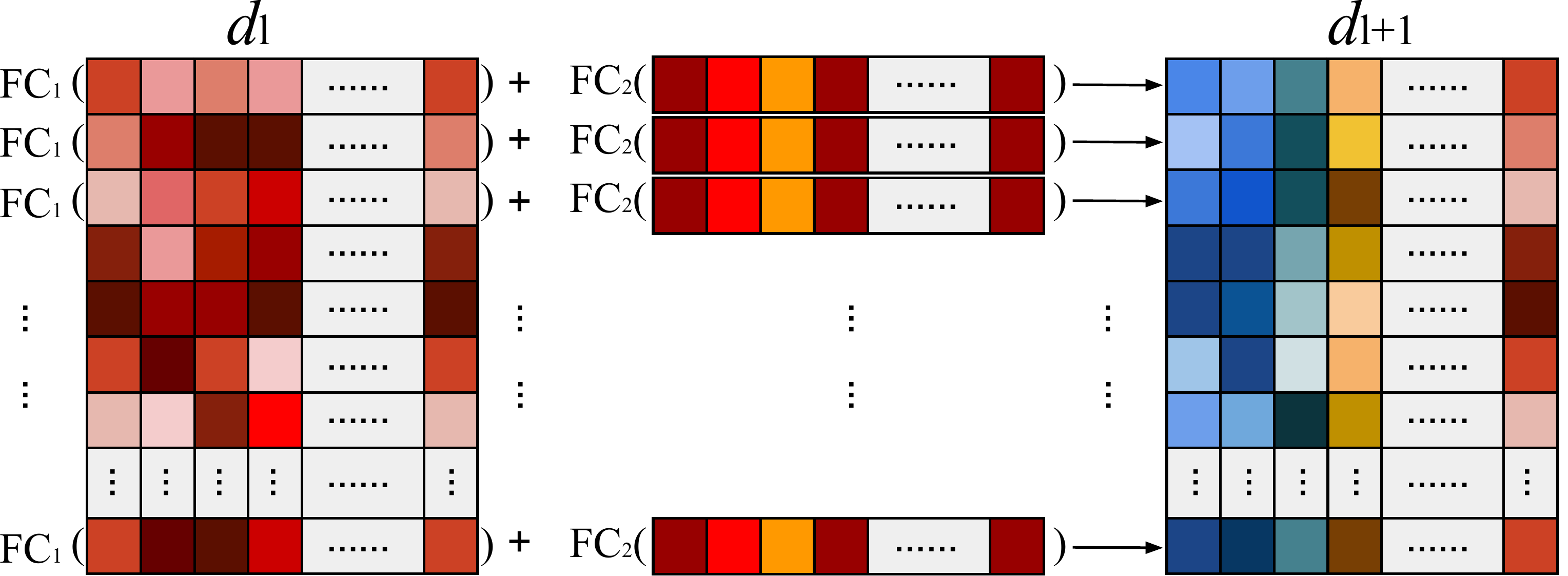}
\end{minipage}
}%
\subfigure[The aggregated feature in (b) and (c).]{
\begin{minipage}[t]{0.36\linewidth}
\centering
\includegraphics[width=\linewidth]{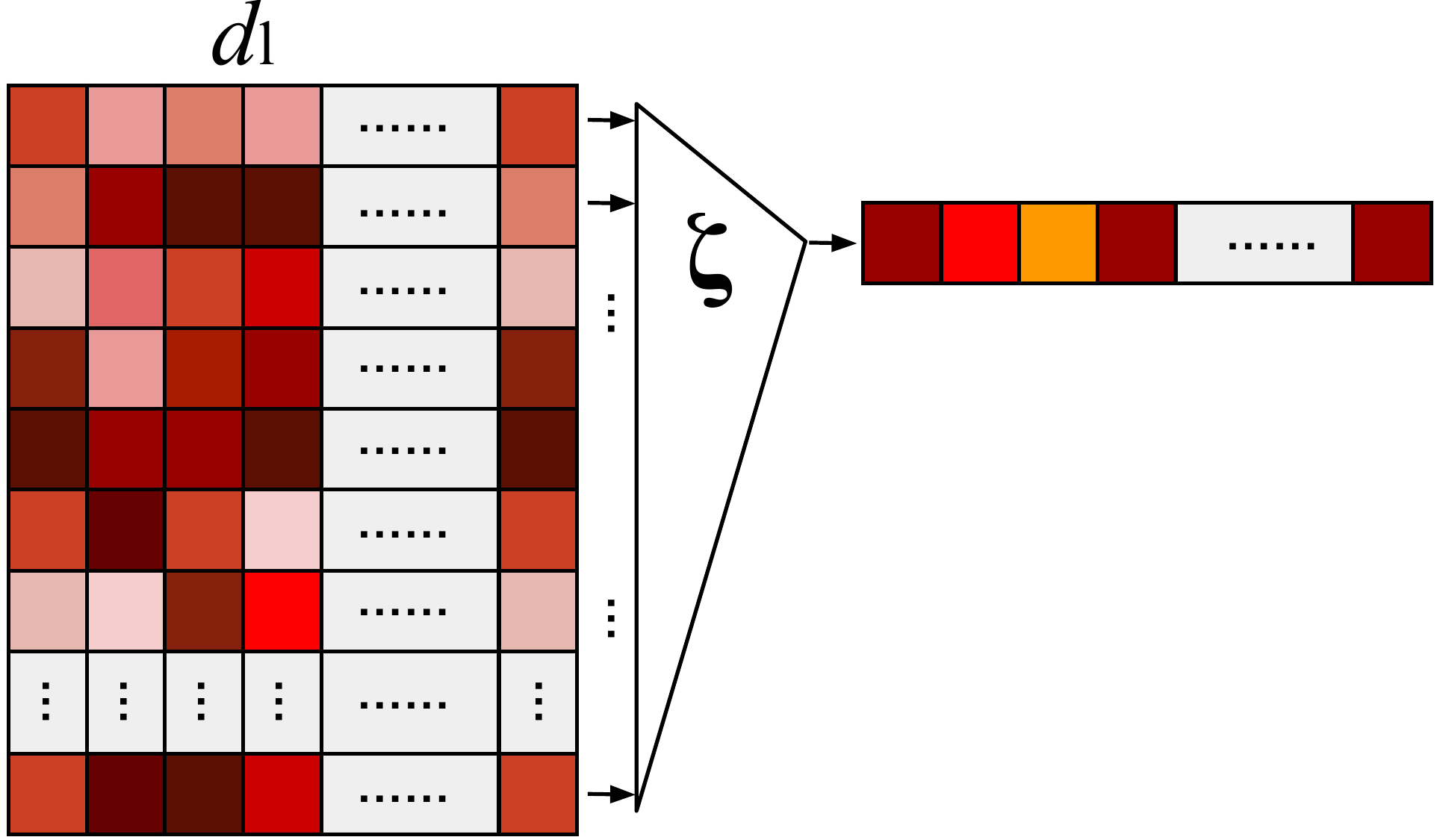}
\end{minipage}
}%

\centering
\vspace{-0.2cm}
\caption{Different architectures of $\phi(\cdot)$ in PointNet, DeepSets, and DSS.}
\label{fig:arch_three}
\end{figure*}

We visualize the differences of $\phi(\cdot)$ in \Figref{fig:arch_three}, and summarize the layer information in Table~\ref{tb:layer_info}.

\begin{table*}[!h]
\caption{Layer information of PointNet, DeepSets, and DSS. BN represents a batch normalization layer.}
\label{tb:layer_info}
\begin{center}
\begin{tabular}{|c|c|c|}
\hline
\bf PointNet & \bf DeepSets & \bf DSS \\
\hline
\hline
$\phi_1: n \times 3 \rightarrow n \times 64$ & $\phi_1:  n \times 3 \rightarrow n \times 256$ & $\phi_1: n \times 3 \rightarrow n \times 64$ \\
BN + ReLU & BN + ELU & BN + ReLU\\
$\phi_2: n \times 64 \rightarrow n \times 64$ & $\phi_2:  n \times 256 \rightarrow n \times 256$ & $\phi_2: n \times 64 \rightarrow n \times 256$ \\
BN + ReLU & BN + ELU & BN + ReLU\\
$\phi_3: n \times 64 \rightarrow n \times 128$ & $\rho: n \times 256 \rightarrow 256$ & $\phi_3: n \times 256 \rightarrow n \times 256$ \\
BN + ReLU & $\sigma_1: 256 \rightarrow 256$ & BN + ReLU\\
$\phi_4: n \times 128 \rightarrow n \times 1024$ & BN + Tanh & $\rho: n \times 256 \rightarrow 256$\\
BN + ReLU & $\sigma_2: 256 \rightarrow 40$ & $\sigma_1: 256 \rightarrow 256$ \\
$\rho: n \times 1024 \rightarrow 1024$ &  & BN + ReLU \\
$\sigma_1: 1024 \rightarrow 512$ & & $\sigma_2: 256 \rightarrow 40$ \\
BN + ReLU &  & \\
$\sigma_2: 512 \rightarrow 256$ &  &  \\
BN + ReLU & & \\
$\sigma_3: 256 \rightarrow 40$ & &  \\
\hline
\end{tabular}
\end{center}
\end{table*}

\subsection{Parametric Pooling Design And Implementation}
\label{ap:para_pool}

We have introduced \verb+ATT+ in Section 4.3.1. In our implementation, we choose $L=512$ so that $\mV \in \R^{512 \times 1024}$ to train the backbone models.

\verb+ATT-GATE+ is a variant of \verb+ATT+ with more learnable parameters:
\begin{equation}
\begin{split}
    \vg & = \sum_{i=1}^{n} a_i \cdot \vf_i \\
    a_i & = \frac{\exp(\vw^\top \cdot (\tanh(\mV \cdot {\vf}_i^\top) \odot \mathrm{sigm}(\mU \cdot {\vf}_i^\top)))}{\sum_{j=1}^{n} \exp(\vw^\top \cdot (\tanh(\mV \cdot {\vf}_j^\top) \odot \mathrm{sigm}(\mU \cdot {\vf}_j^\top)))}
\end{split}
\end{equation}
where $\mU, \mV \in \R^{L \times M}$, $\mathrm{sigm}(\cdot)$ is the sigmoid activation function, and $\odot$ is an element-wise multiplication. We also choose $L = 512$ in \verb+ATT-GATE+ to train the backbones.

\verb+PMA+~\cite{lee2019set} adopts multi-head attention into pooling on a learnable set of $k$ seed vectors $\mS \in \R^{k \times d_m}$ Let $\mF \in \R^{n \times d_m}$ be the matrix version of the set of features. 
\begin{equation}
\begin{split}
    & \mathrm{PMA}_k(\mF) = \mathrm{MAB}(\mS,\mathrm{FC}(\mF)) \\
    & \mathrm{MAB}(\mX, \mY) = \mH + \mathrm{FC}(\mH) \\
    & \mathrm{where} \quad \mH = \mX + \mathrm{Multihead}(\mX,\mY,\mY;\vw)
\end{split}
\end{equation}

where $\mathrm{FC}(\cdot)$ is the fully connected layer and $\mathrm{Multihead}(\cdot)$ is the multi-head attention module~\cite{vaswani2017attention}. We follow the implementation in the released codebase\footnote{\url{https://github.com/juho-lee/set_transformer}} to choose $k=1$, the number of head = 4, and the hidden neurons in $\mathrm{FC}(\cdot)$ = 128 to train the backbone models.

Since \verb+SoftPool+~\cite{wang2020softpoolnet} sorts the feature set in each dimension, it requires the number of dimensions $d_m$ to be relatively small. We follow the description in their paper to choose $d_m = 8$ and $k = 32$ so that each ${\mF_j}' \in \R^{32 \times 8}$. We apply one convolutional layer to aggregate each ${\mF'}_j$ into $g_j \in \R^{1 \times 32}$ so that the final $\vg \in \R^{1 \times 256}$. Therefore, for all backbone networks with \verb+SoftPool+, we apply the last equivariant layer as $\phi: n \times d_{m-1} \rightarrow n \times 8$ and $\rho: n \times 8 \rightarrow 256$.
\section{\texttt{DeepSym} Ablations}
\label{ap:deepsym}

It is worth noting that \verb+DeepSym+ does not require the final layer to have only one neuron. However, to have a fair comparison with other pooling operations that aggregate into one feature from each dimension, our implementation of \verb+DeepSym+ also aggregates into one feature from each dimension.

\subsection{Evaluation Details}
\label{ap:deepsym_eval}

\begin{figure}[!h]
\begin{center}
\includegraphics[width=\linewidth]{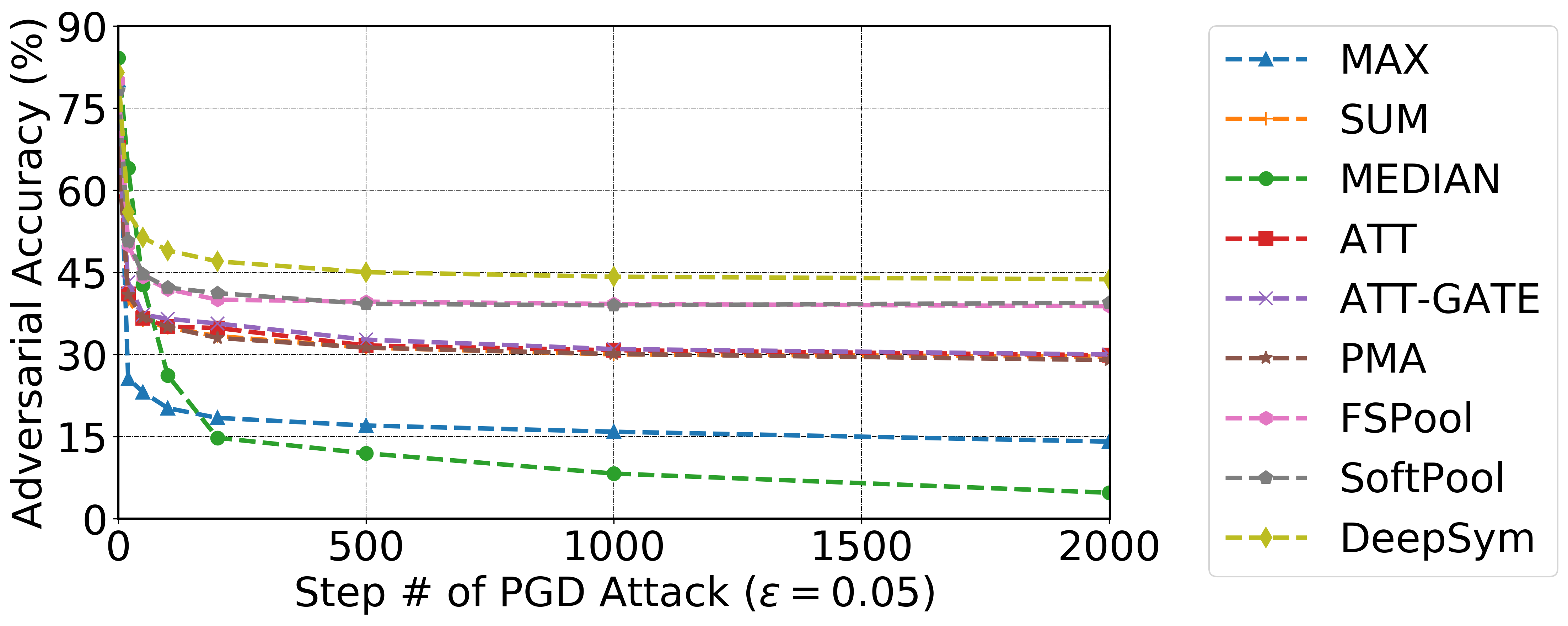} 
\end{center}
\caption{Adversarial accuracy of PGD-20 trained PointNet with different pooling operations.  We leverage the PGD attack with different steps to evaluate the model's robustness.}
\label{fig:2000}
\end{figure}

\begin{figure}[!h]
\begin{center}
\includegraphics[width=\linewidth]{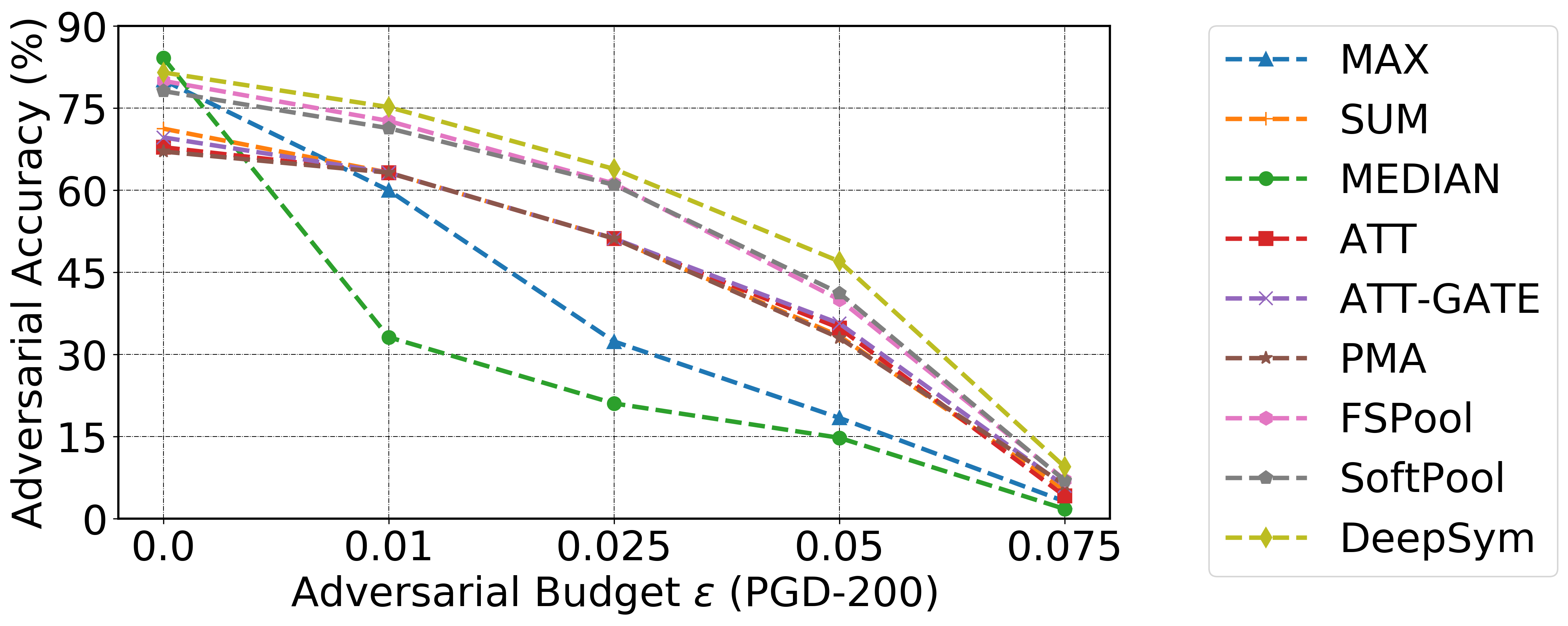} 
\end{center}
\caption{Adversarial accuracy of PGD-20 trained PointNet with different pooling operations. We leverage the PGD attack with different budgets to evaluate the model's robustness.}
\label{fig:epsilon}
\end{figure}

We also perform extensive evaluations using different PGD attack steps and budgets $\epsilon$ on PGD-20 trained PointNet. \Figref{fig:2000} shows that PointNet with \verb+DeepSym+ consistently achieves the best adversarial accuracy. We also validate \verb+MEDIAN+ pooling indeed hinders the gradient backward propagation. The adversarial accuracy of PointNet with \verb+MEDIAN+ pooling consistently drops even after PGD-1000. However, the adversarial accuracy of PointNet with other pooling operations usually converges after PGD-200. \Figref{fig:epsilon} shows that \verb+DeepSym+ also outperforms other pooling operations under different adversarial budgets $\epsilon$.

We leverage the default setup in FGSM, BIM, and MIM in our evaluation. FGSM is a single-step attack method, which can be represented as:
\begin{equation}
    \mX_{adv} = \mX + \epsilon \cdot \sign(\nabla_{\mX} \mathcal{L}(\mX,\vtheta,\vy))
    \label{eq:pgd}
\end{equation}
The BIM attack is similar to PGD attacks described in Appendix~\ref{ap:adap_setup}. The differences are 1) the attack starts from the original point cloud $\mX$ and 2) the step size $\alpha = \epsilon / T$, where $T$ is the number of attack steps. The MIM attack introduces momentum terms into the adversarial optimization:
\begin{equation}
    \vg_{t+1} = \mu \cdot \vg_{t} +  \frac{\nabla_{\mX_{t}} \mathcal{L}(\mX_{t},\vtheta,\vy)}{||\nabla_{\mX_{t}} \mathcal{L}(\mX_{t},\vtheta,\vy))||_{1}}
    \label{eq:pgd}
\end{equation}
\begin{equation}
    \mX_{t+1} = \mX_{t} + \alpha \cdot \sign(\vg_{t+1})
    \label{eq:pgd}
\end{equation}
Similar to BIM, the attack starts from the original point cloud $\mX$ and the step size $\alpha = \epsilon / T$. We set $\mu = 1$ following the original setup~\cite{dong2018boosting}.

Due to the computational resource constraints, we set the sample size = 32 and allow 2000 quires to find each adversarial example in the score-based black-box attack~\cite{uesato2018adversarial,ilyas2018black}. For the evolution attack, we use the default loss $\mathcal{L}$ as the fitness score, and initialize 32 sets of perturbations from a Gaussian distribution $\mathcal{N}(0,1)$. 4 sets of perturbations with top fitness scores will remain for the next iteration, while others will be discarded. We also allow 2000 generations of evolution to find the adversarial example.

\subsection{Evaluation on ScanObjectNN}
\label{ap:scanobj}

We also evaluate the adversarial robustness of different pooling operations on a new point cloud dataset, ScanObjectNN~\cite{uy-scanobjectnn-iccv19}, which contains 2902 objects belonging to 15 categories. We leverage the same adversarial training setup as ModelNet10 (\textit{i.e.,} PGD-1). Table~\ref{tb:scan} shows the results. We find that PointNet with \verb+DeepSym+ still achieves the best adversarial robustness. Since the point clouds from ScanObjectNN are collected from real-world scenes, which suffers from occlusion and imperfection, both nominal and adversarial accuracy drops compared to the results ModelNet40. We find that even some clean point clouds cannot be correctly recognized by human perception. Therefore, the performance degradation is also expected and we believe the results are not as representative as ones on ModelNet40.

\begin{table}[!h]
\footnotesize
\vspace{-0.2cm}
\caption{Adversarial robustness of PointNet with different pooling operations under PGD-200 at $\epsilon=0.05$.}
\vspace{-0.2cm}
\label{tb:scan}
\begin{center}
\begin{tabular}{|c|c|c|}
\hline
\makecell*[c]{\bf Pooling Operation} &\multicolumn{1}{c|}{\bf Nominal Accuracy} &\multicolumn{1}{c|}{\bf Adversarial Accuracy} \\
\hline
\hline
\verb+MAX+ &75.2\%   &16.8\% \\
\hline
\verb+MEDIAN+ &68.4\%   &8.2\% \\
\hline
\verb+SUM+ &63.5\%   &18.3\% \\
\hline
\verb+ATT+ &62.7\%   &17.9\% \\
\hline
\verb+ATT-GATE+  &59.8\%   &17.1\%  \\
\hline
\verb+PMA+   &61.2\%   &16.2\% \\
\hline
\verb+FSPool+   &\textbf{76.8\%}   &20.1\%\\
\hline
\verb+SoftPool+   &73.2\%   &17.2\% \\
\hline
\verb+DeepSym+ (ours)   &76.7\%   &\textbf{22.8\%} \\
\hline
\end{tabular}
\end{center}
\vspace{-0.4cm}
\end{table}

\subsection{T-SNE Visualizations}
\begin{figure*}[t]
\begin{minipage}[t]{0.85\linewidth}
\subfigure[\texttt{MAX} pooling on training data, validation data, and PGD-200 adversarial validation data.]{\begin{minipage}[t]{0.32\linewidth}
\vspace{0.0cm}
\begin{center}
\includegraphics[width=\linewidth]{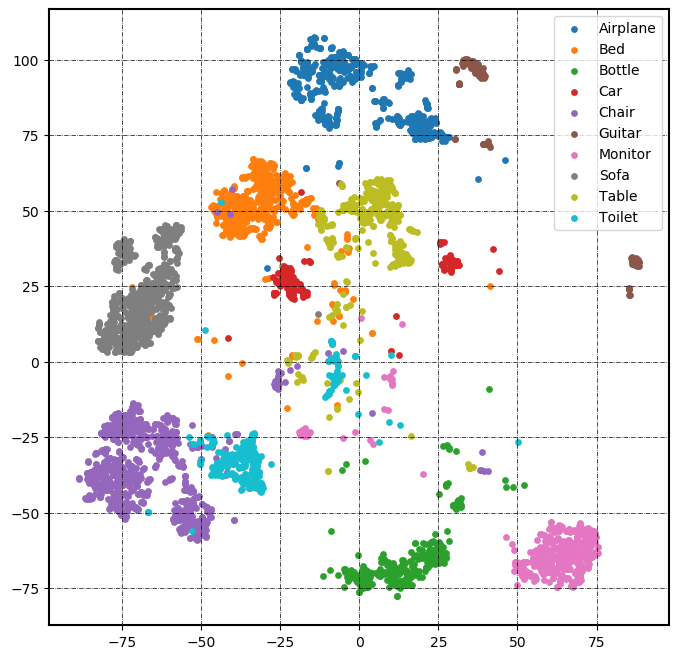} 
\end{center}
\end{minipage}
\hfill
\begin{minipage}[t]{0.32\linewidth}
\vspace{0.0cm}
\begin{center}
\includegraphics[width=\linewidth]{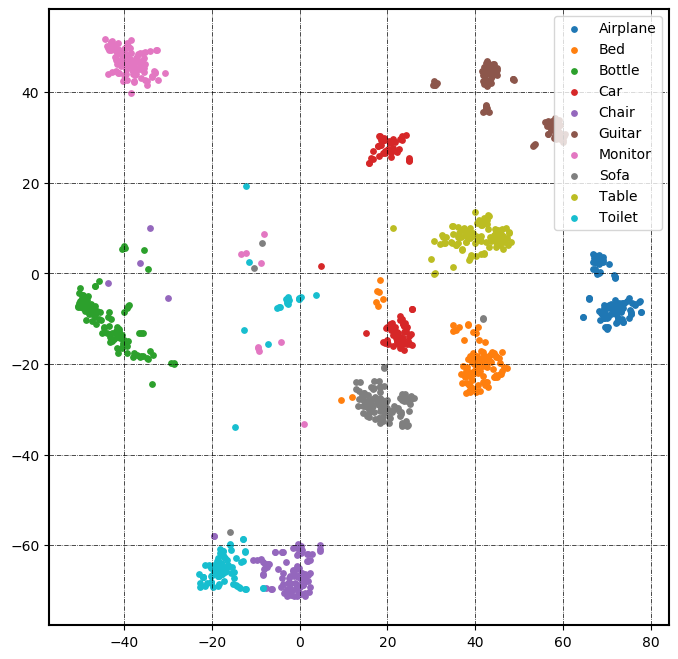} 
\end{center}
\end{minipage}
\hfill
\begin{minipage}[t]{0.32\linewidth}
\vspace{0.0cm}
\begin{center}
\includegraphics[width=\linewidth]{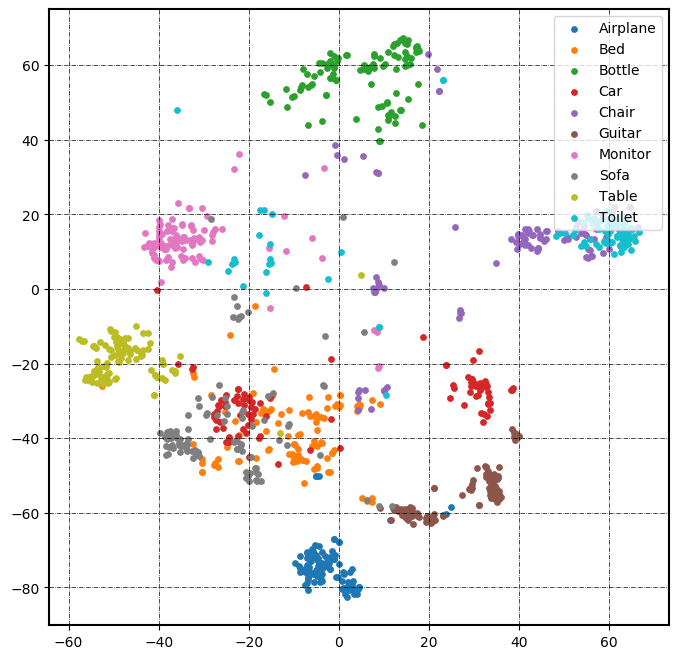} 
\end{center}
\end{minipage}}

\subfigure[\texttt{FSPool} on training data, validation data, and PGD-200 adversarial validation data.]{\begin{minipage}[t]{0.32\linewidth}
\vspace{0.0cm}
\begin{center}
\includegraphics[width=\linewidth]{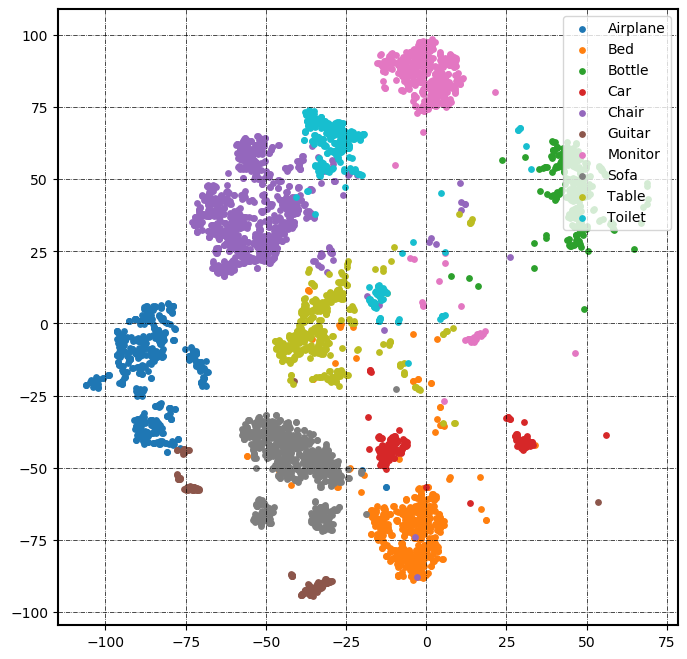} 
\end{center}
\end{minipage}
\hfill
\begin{minipage}[t]{0.32\linewidth}
\vspace{0.0cm}
\begin{center}
\includegraphics[width=\linewidth]{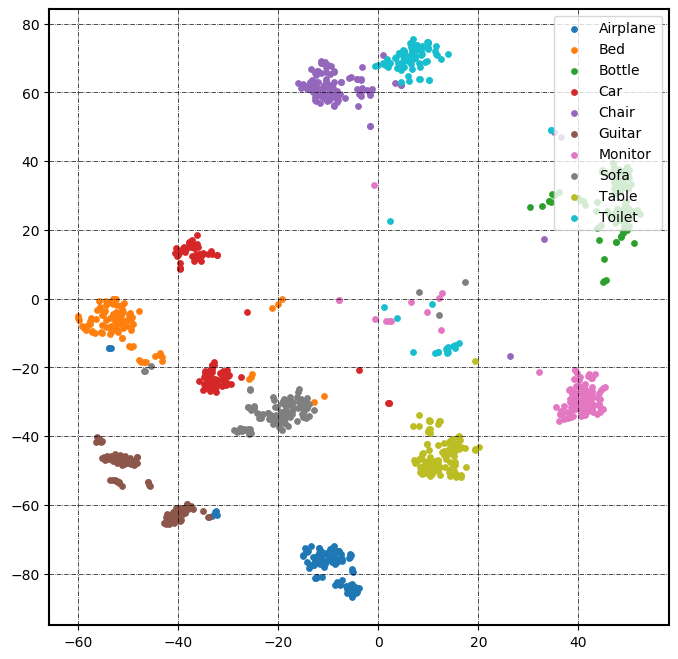} 
\end{center}
\end{minipage}
\hfill
\begin{minipage}[t]{0.32\linewidth}
\vspace{0.0cm}
\begin{center}
\includegraphics[width=\linewidth]{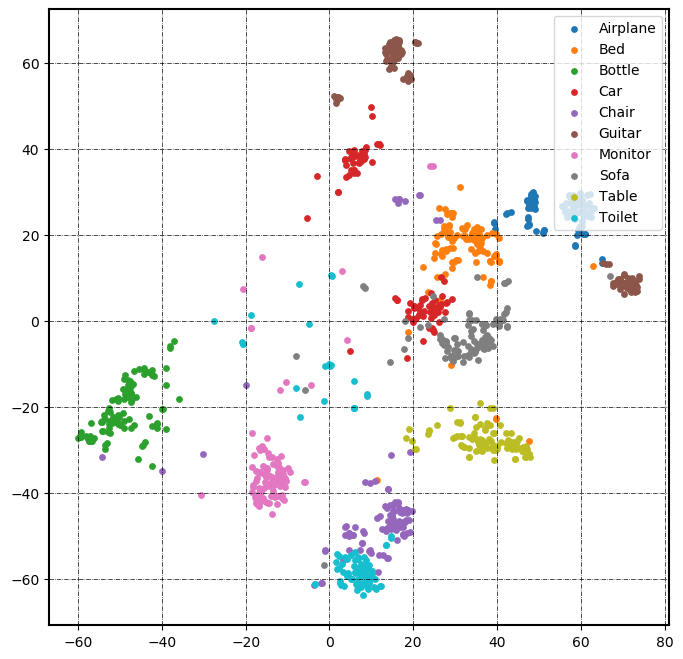} 
\end{center}
\end{minipage}}

\subfigure[\texttt{SoftPool} on training data, validation data, and PGD-200 adversarial validation data.]{\begin{minipage}[t]{0.32\linewidth}
\vspace{0.0cm}
\begin{center}
\includegraphics[width=\linewidth]{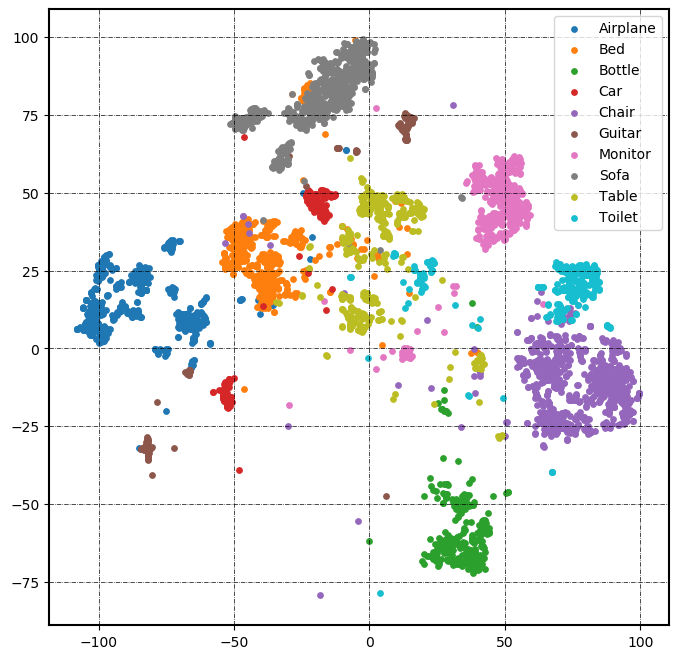} 
\end{center}
\end{minipage}
\hfill
\begin{minipage}[t]{0.32\linewidth}
\vspace{0.0cm}
\begin{center}
\includegraphics[width=\linewidth]{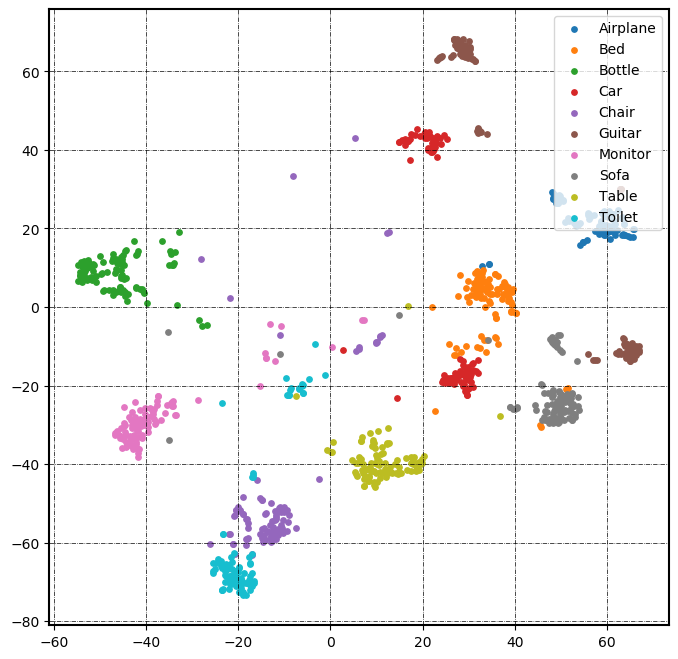} 
\end{center}
\end{minipage}
\hfill
\begin{minipage}[t]{0.32\linewidth}
\vspace{0.0cm}
\begin{center}
\includegraphics[width=\linewidth]{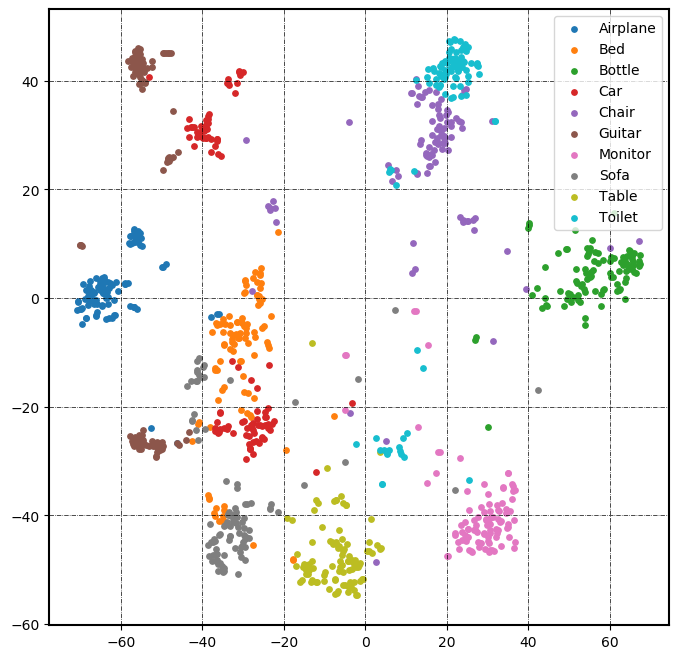} 
\end{center}
\end{minipage}}


\subfigure[\texttt{DeepSym} on training data, validation data, and PGD-200 adversarial validation data.]{\begin{minipage}[t]{0.32\linewidth}
\vspace{-0.02cm}
\begin{center}
\includegraphics[width=\linewidth]{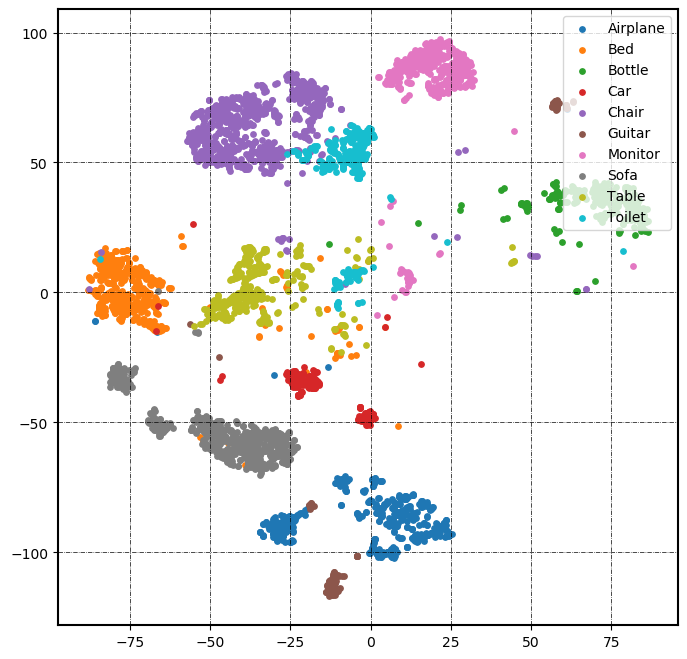} 
\end{center}
\end{minipage}
\hfill
\begin{minipage}[t]{0.32\linewidth}
\vspace{0.0cm}
\begin{center}
\includegraphics[width=\linewidth]{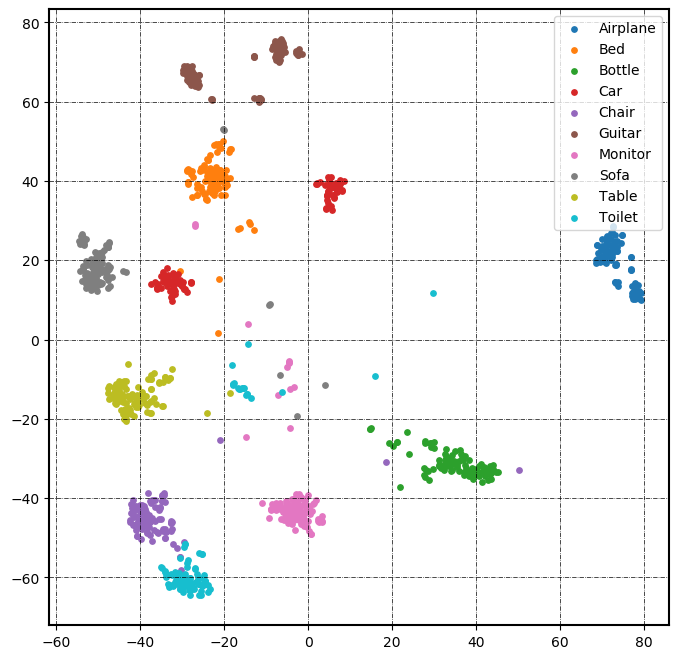} 
\end{center}
\end{minipage}
\hfill
\begin{minipage}[t]{0.32\linewidth}
\vspace{0.0cm}
\begin{center}
\includegraphics[width=\linewidth]{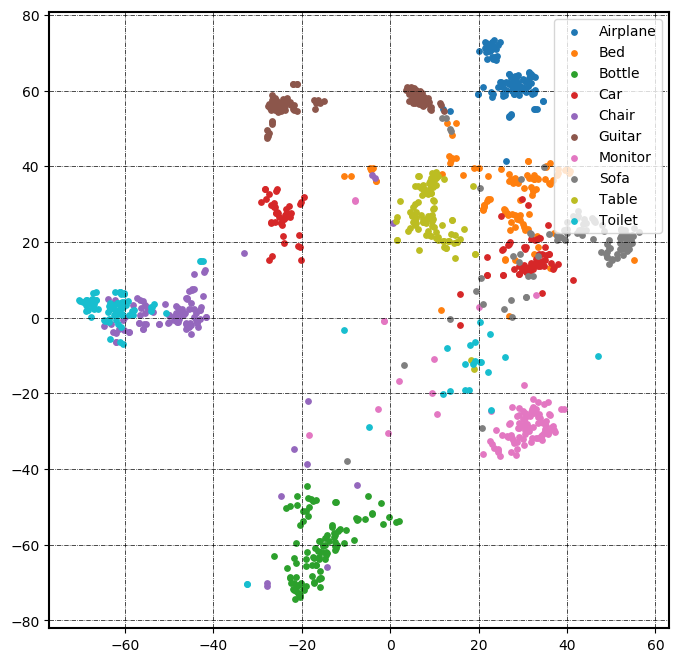} 
\end{center}
\end{minipage}}
\vspace{-0.15cm}
\caption{T-SNE visualizations of PointNet \underline{feature embeddings} with \texttt{MAX}, \texttt{FSPool}, \texttt{SoftPool}, and \texttt{DeepSym} pooling operations. Three columns correspond to training data, validation data, and PGD-200 adversarial validation data, from left to right.}
\label{fig:tsne_feature}
\end{minipage}
\end{figure*}

\begin{figure*}[t]
\begin{minipage}[t]{0.85\linewidth}
\subfigure[\texttt{MAX} pooling on training data, validation data, and PGD-200 adversarial validation data.]{\begin{minipage}[t]{0.32\linewidth}
\vspace{0.0cm}
\begin{center}
\includegraphics[width=\linewidth]{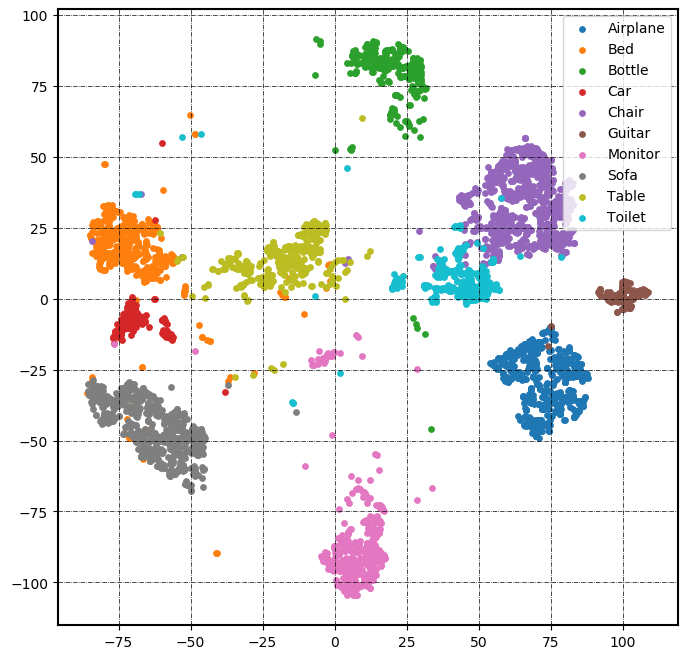} 
\end{center}
\end{minipage}
\hfill
\begin{minipage}[t]{0.32\linewidth}
\vspace{0.0cm}
\begin{center}
\includegraphics[width=\linewidth]{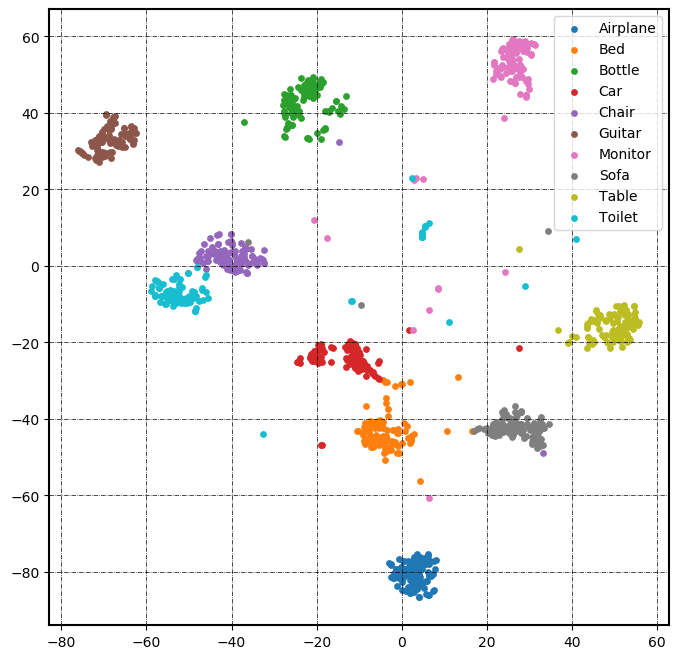} 
\end{center}
\end{minipage}
\hfill
\begin{minipage}[t]{0.32\linewidth}
\vspace{0.0cm}
\begin{center}
\includegraphics[width=\linewidth]{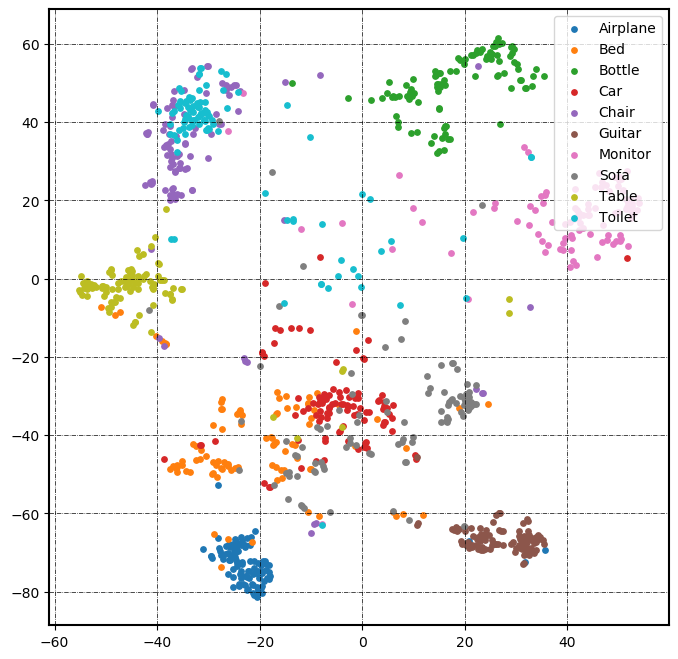} 
\end{center}
\end{minipage}}

\subfigure[\texttt{FSPool} on training data, validation data, and PGD-200 adversarial validation data.]{\begin{minipage}[t]{0.32\linewidth}
\vspace{0.0cm}
\begin{center}
\includegraphics[width=\linewidth]{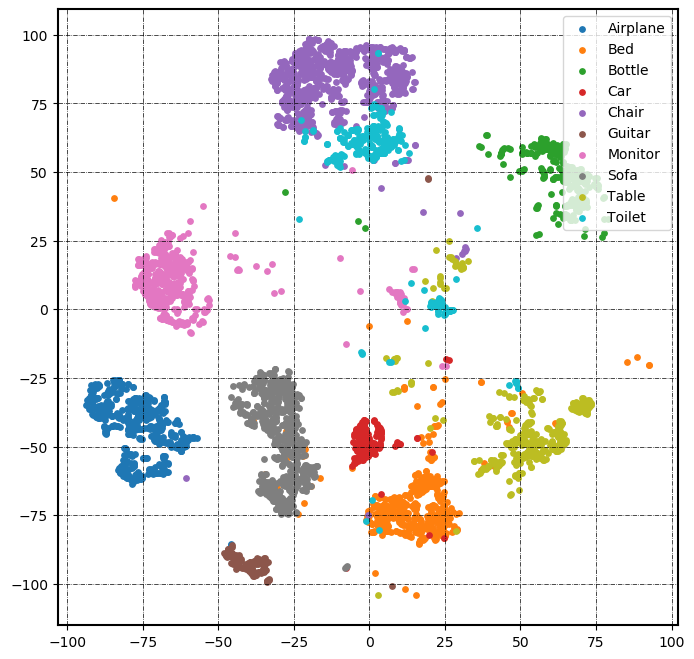} 
\end{center}
\end{minipage}
\hfill
\begin{minipage}[t]{0.32\linewidth}
\vspace{0.0cm}
\begin{center}
\includegraphics[width=\linewidth]{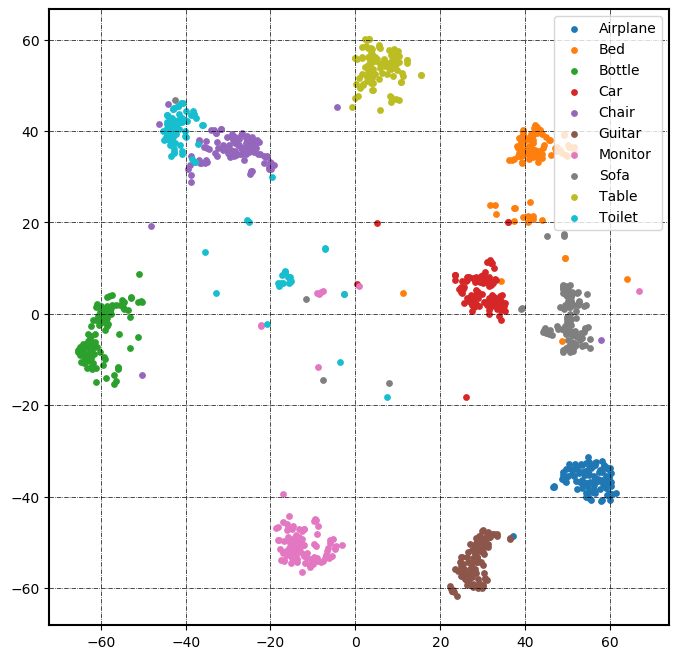} 
\end{center}
\end{minipage}
\hfill
\begin{minipage}[t]{0.32\linewidth}
\vspace{0.0cm}
\begin{center}
\includegraphics[width=\linewidth]{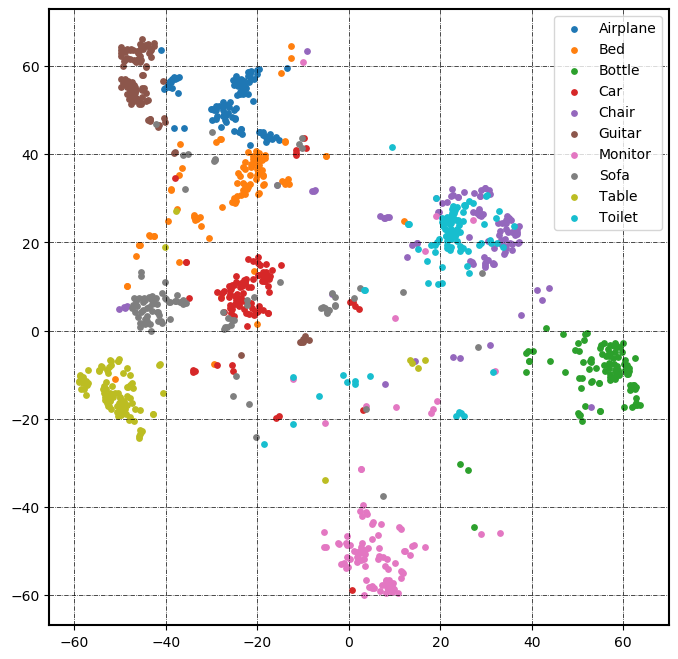} 
\end{center}
\end{minipage}}

\subfigure[\texttt{SoftPool} on training data, validation data, and PGD-200 adversarial validation data.]{\begin{minipage}[t]{0.32\linewidth}
\vspace{0.0cm}
\begin{center}
\includegraphics[width=\linewidth]{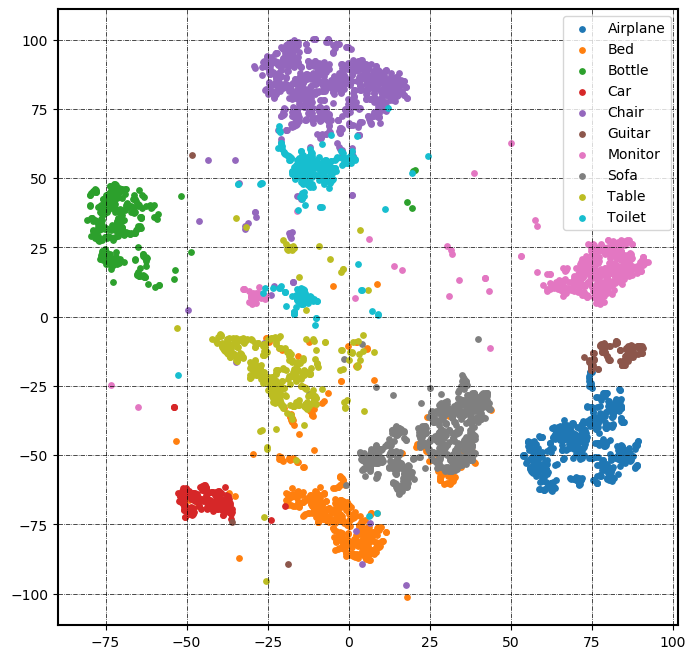} 
\end{center}
\end{minipage}
\hfill
\begin{minipage}[t]{0.32\linewidth}
\vspace{0.0cm}
\begin{center}
\includegraphics[width=\linewidth]{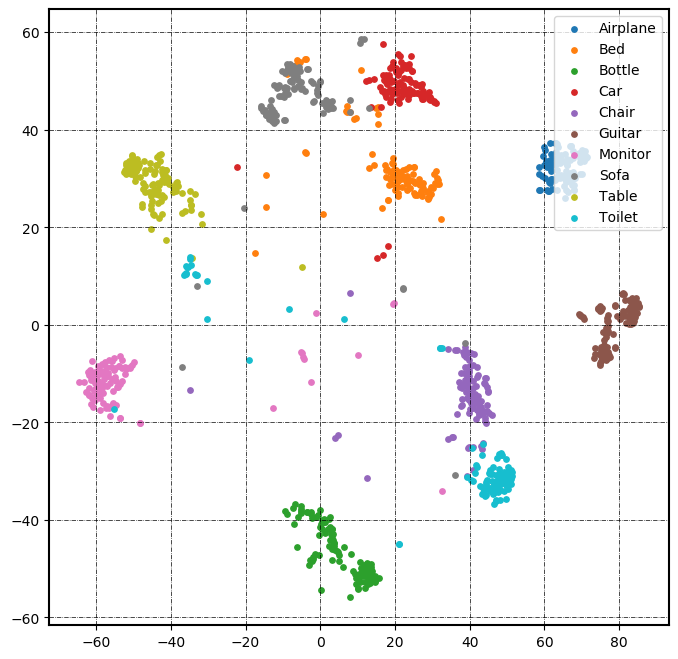} 
\end{center}
\end{minipage}
\hfill
\begin{minipage}[t]{0.32\linewidth}
\vspace{0.0cm}
\begin{center}
\includegraphics[width=\linewidth]{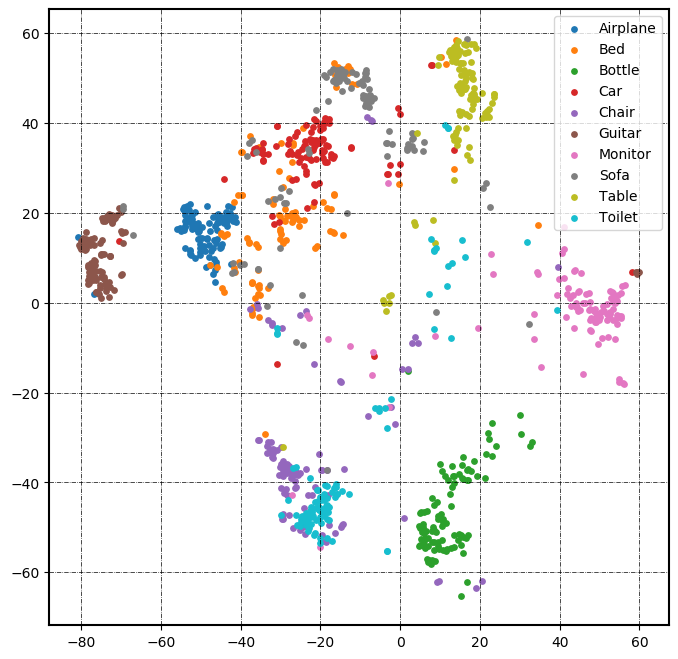} 
\end{center}
\end{minipage}}


\subfigure[\texttt{DeepSym} on training data, validation data, and PGD-200 adversarial validation data.]{\begin{minipage}[t]{0.32\linewidth}
\vspace{-0.02cm}
\begin{center}
\includegraphics[width=\linewidth]{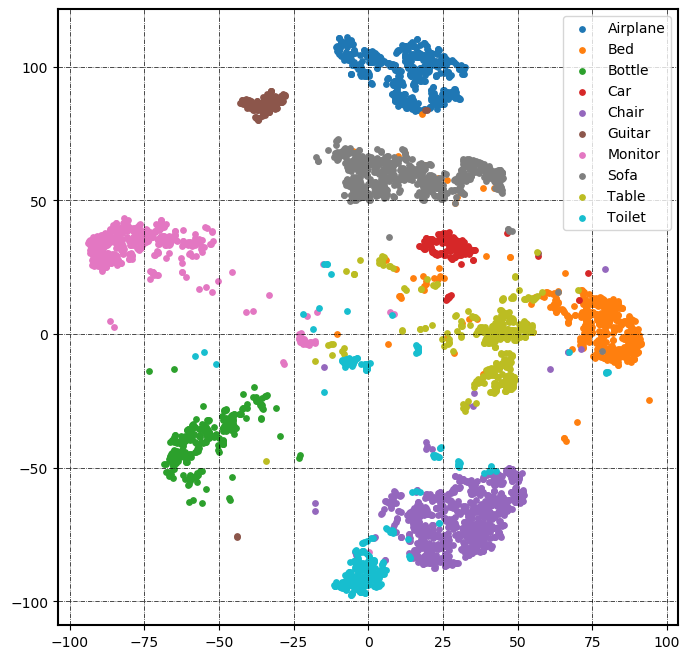} 
\end{center}
\end{minipage}
\hfill
\begin{minipage}[t]{0.32\linewidth}
\vspace{0.0cm}
\begin{center}
\includegraphics[width=\linewidth]{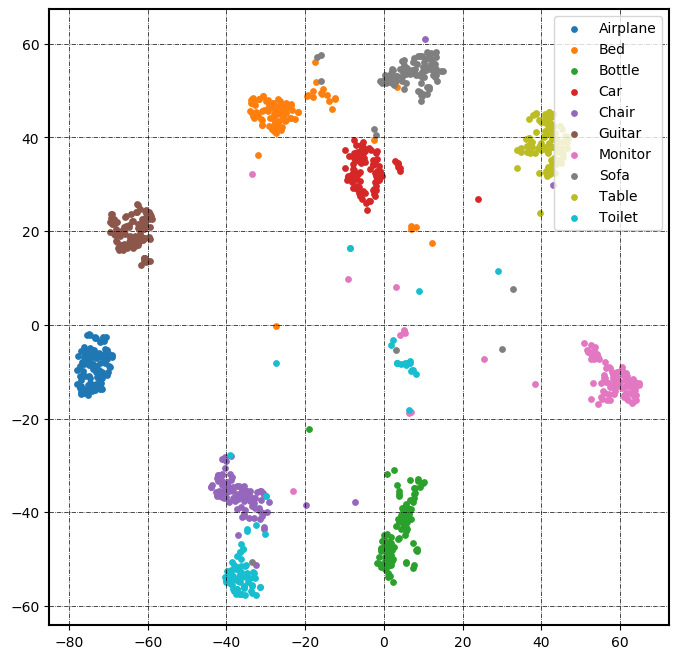} 
\end{center}
\end{minipage}
\hfill
\begin{minipage}[t]{0.32\linewidth}
\vspace{0.0cm}
\begin{center}
\includegraphics[width=\linewidth]{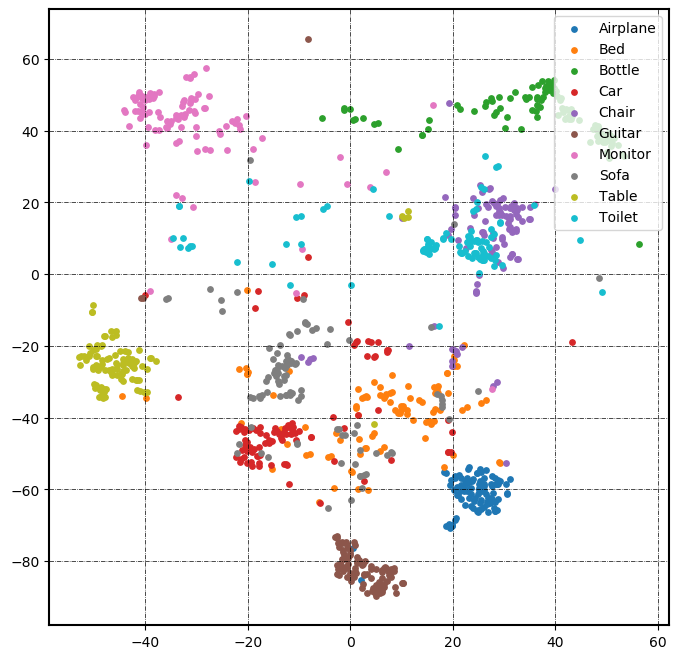} 
\end{center}
\end{minipage}}
\vspace{-0.15cm}
\caption{T-SNE visualizations of PointNet \underline{logits} with \texttt{MAX}, \texttt{FSPool}, \texttt{SoftPool}, and \texttt{DeepSym} pooling operations. Three columns correspond to training data, validation data, and PGD-200 adversarial validation data, from left to right.}
\label{fig:tsne_logits}
\end{minipage}
\end{figure*}

We visualize the global feature embeddings of adversarially trained PointNet under PGD-20 with different pooling operations in \Figref{fig:tsne_feature} and their logits in \Figref{fig:tsne_logits}. Since it is hard to pick 40 distinct colors, though we put all data from 40 classes into the T-SNE process, we only choose 10 categories from ModelNet40 to realize the visualizations. 

\end{document}